\documentclass{article}

\usepackage[preprint]{neurips_2024}

\usepackage{amssymb}            % Defines common symbols like \mathbb R
\usepackage{mathtools}          % Extends amsmath, providing common math tools
\usepackage{mathrsfs}           % Enables \mathscr, which can work in cases that \mathcal does not
\usepackage{graphicx}           % For including images
\usepackage{xcolor}
\usepackage{physics}           % For including images
\usepackage{subcaption}         % Allows for the use of subfigures and subcaptions
\usepackage[space]{grffile}     % For spaces in image names
\usepackage{url}                % For displaying urls
\usepackage[subtle,title=normal,mathspacing=normal,paragraphs=normal,wordspacing=normal,floats=normal]{savetrees}
\usepackage{comment}
\usepackage{floatrow}
\usepackage{algorithm2e}
\usepackage{wrapfig}  % For wrapping text around an algorithm
\usepackage{booktabs} % For line spacing in tables
\usepackage{array}    % To change spacing between columns with a multicol
\usepackage{tikz}
\usetikzlibrary{patterns}
\usepackage[most]{tcolorbox}  % For hatching the background of a text 
\tcbuselibrary{fitting}

\usepackage{titlesec}

\titlespacing{\paragraph}{%
  0pt}{%              left margin
  0\baselineskip}{% space before (vertical)
  0.5em}%  

\let\phi=\varphi
\let\epsilon=\varepsilon

\newcommand{\tp}{\intercal}
\newcommand{\ssp}{\mathcal{S}}
\newcommand{\A}{\mathcal{A}}
\newcommand{\rw}{\mathbf{r}}
\newcommand{\R}{\mathbb{R}}
\newcommand{\E}{\mathbb{E}}
\newcommand{\vpi}{\mathbf{V}^\pi}
\newcommand{\qpi}{\mathbf{Q}^\pi}
\newcommand{\api}{\mathbf{A}^\pi}
\newcommand{\hq}{\widehat{\mathbf{Q}}} % reward-to-go
\newcommand{\vt}{\mathbf{V}^\theta}
\newcommand{\vc}{\widetilde{\mathbf{V}}_\psi}
\newcommand{\w}{\mathbf{w}}
\newcommand{\p}{\mathbf{p}}
\newcommand{\q}{\mathbf{q}}
\newcommand{\z}{\mathbf{z}}
\newcommand{\g}{\mathbf{g}}
\newcommand{\balpha}{\boldsymbol{\alpha}}
\newcommand{\bmu}{\boldsymbol{\mu}}
\newcommand{\bsigma}{\boldsymbol{\sigma}}
\newcommand{\hA}{\skew{6}{\hat}{A}}

\newcommand{\hnab}{\widehat{\nabla}}
\newcommand{\dk}{\Delta_K}
\newcommand{\J}{\mathbf{J}}
\newcommand{\D}{\mathcal{D}}
\newcommand{\bod}{\mathbf{f}_\zeta}

\title{In Search for Architectures and Loss Functions in Multi-Objective Reinforcement Learning}
\author{%
    Mikhail Terekhov  \\
    CLAIRE, EPFL \\
    \texttt{mikhail.terekhov@epfl.ch} \\
  \And
    Caglar Gulcehre  \\
    CLAIRE, EPFL \\
    \texttt{caglar.gulcehre@epfl.ch}
}

\begin{document}

\maketitle

\begin{abstract}
Multi-objective reinforcement learning (MORL) is essential for addressing the intricacies of real-world RL problems, which often require trade-offs between multiple utility functions. However, MORL is challenging due to unstable learning dynamics with deep learning-based function approximators. The research path most taken has been to explore different \textbf{value}-based loss functions for MORL to overcome this issue. Our work empirically explores model-free \textbf{policy} learning loss functions and the impact of different architectural choices. We introduce two different approaches: \textit{Multi-objective Proximal Policy Optimization} (MOPPO), which extends PPO to MORL, and \textit{Multi-objective Advantage Actor Critic} (MOA2C), which acts as a simple baseline in our ablations. Our proposed approach is straightforward to implement, requiring only small modifications at the level of function approximator. We conduct comprehensive evaluations on the MORL \emph{Deep Sea Treasure}, \emph{Minecart}, and \emph{Reacher} environments and show that MOPPO effectively captures the Pareto front. Our extensive ablation studies and empirical analyses reveal the impact of different architectural choices, underscoring the robustness and versatility of MOPPO compared to popular MORL approaches like Pareto Conditioned Networks (PCN) and Envelope Q-learning in terms of MORL metrics, including hypervolume and expected utility.
\end{abstract}
\section{Introduction}
\label{sec:intro}
Many optimization problems in the real world require consideration of multiple conflicting objectives. \cite{liu2022accuracy} provide examples of accuracy versus fairness trade-offs in credit scoring and criminal justice, and \cite{vamplew2021potential} show how to trade performance for safety in intelligent agents. \emph{Multi-objective optimization} is the field that studies these problems formally. It is known as \emph{multi-objective reinforcement learning (MORL)} in the sequential decision-making setting. In MORL, we seek \emph{policies} that maximize the respective objectives. A single policy mapping states to actions is insufficient to satisfy all possible trade-offs between objectives; hence, in MORL, we usually discuss sets of policies covering these trade-offs.

The performance of modern MORL approaches is often measured on toy grid-world or 2D locomotion problems, such as those in MO-Gym by~\cite{Alegre2022bnaic}. At the same time, single-objective RL is already used in many practical applications, such as language modeling~\citep{ouyang2022training}, real-world robot locomotion~\citep{fu2023deep}, and control of scientific equipment~\citep{degrave2022magnetic}. One plausible explanation of this gap is that MORL approaches often explicitly store Pareto-optimal policies or rely on Q-learning. For toy problems, this provides optimal coverage, but this is not scalable, and we noticed that the training can suffer from unstable learning dynamics, especially when different rewards interfere with each other.

\begin{figure}
  \centering
  \begin{subfigure}{0.3325\linewidth}
      \includegraphics[height=0.175\textheight]{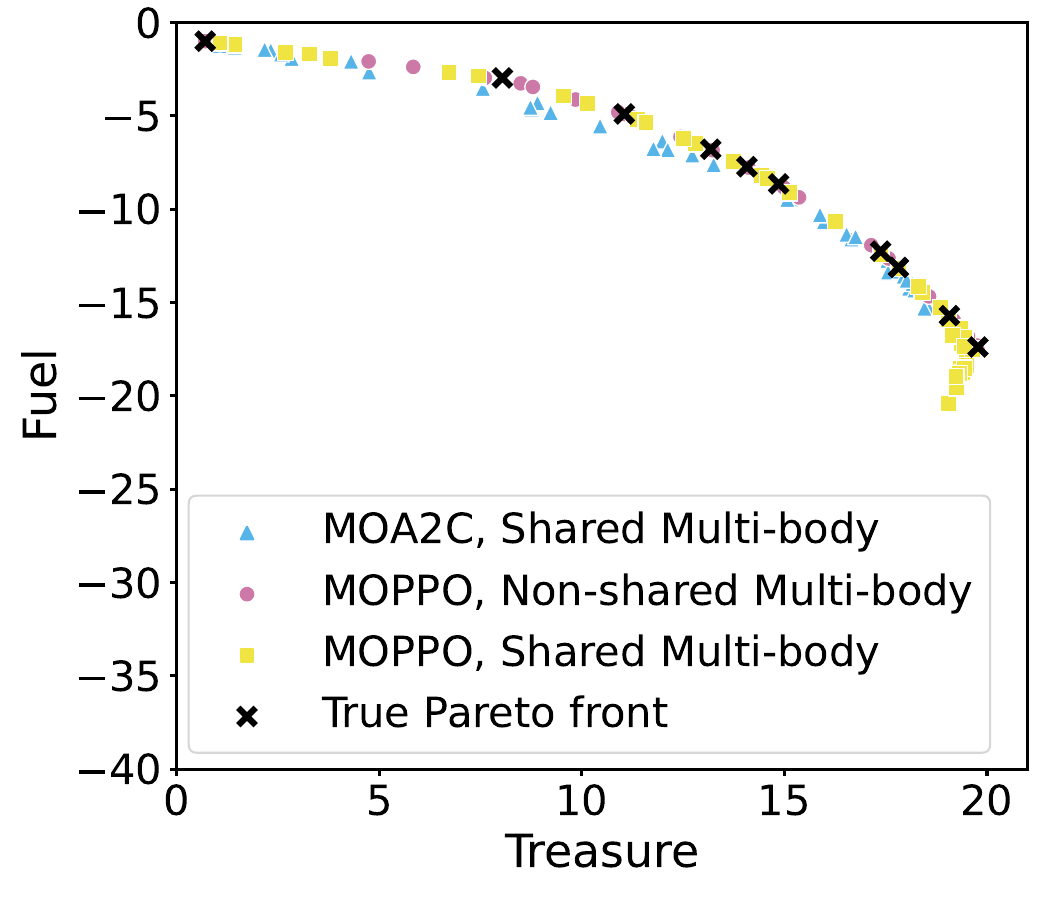}
      \caption{Multi-body architectures}
  \end{subfigure}
  \begin{subfigure}{0.30875\linewidth}
      \includegraphics[height=0.175\textheight]{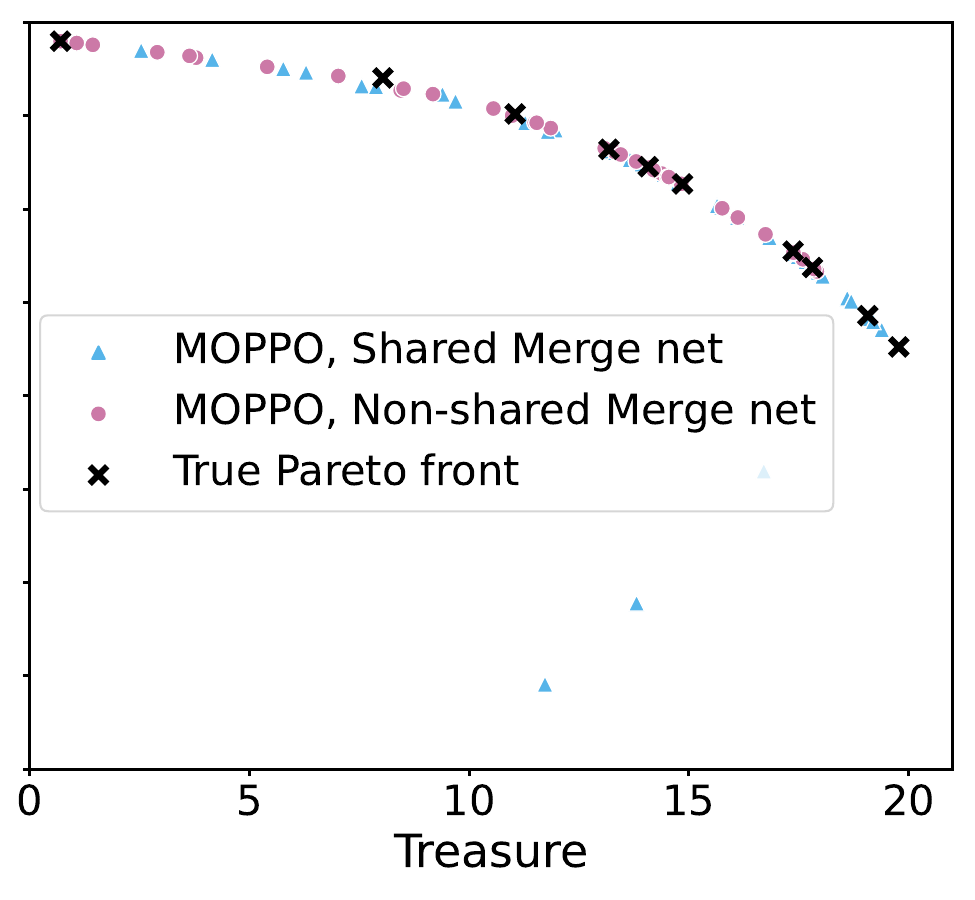}
      \caption{Merge net architectures}
  \end{subfigure}
  \begin{subfigure}{0.30875\linewidth}
      \includegraphics[height=0.175\textheight]{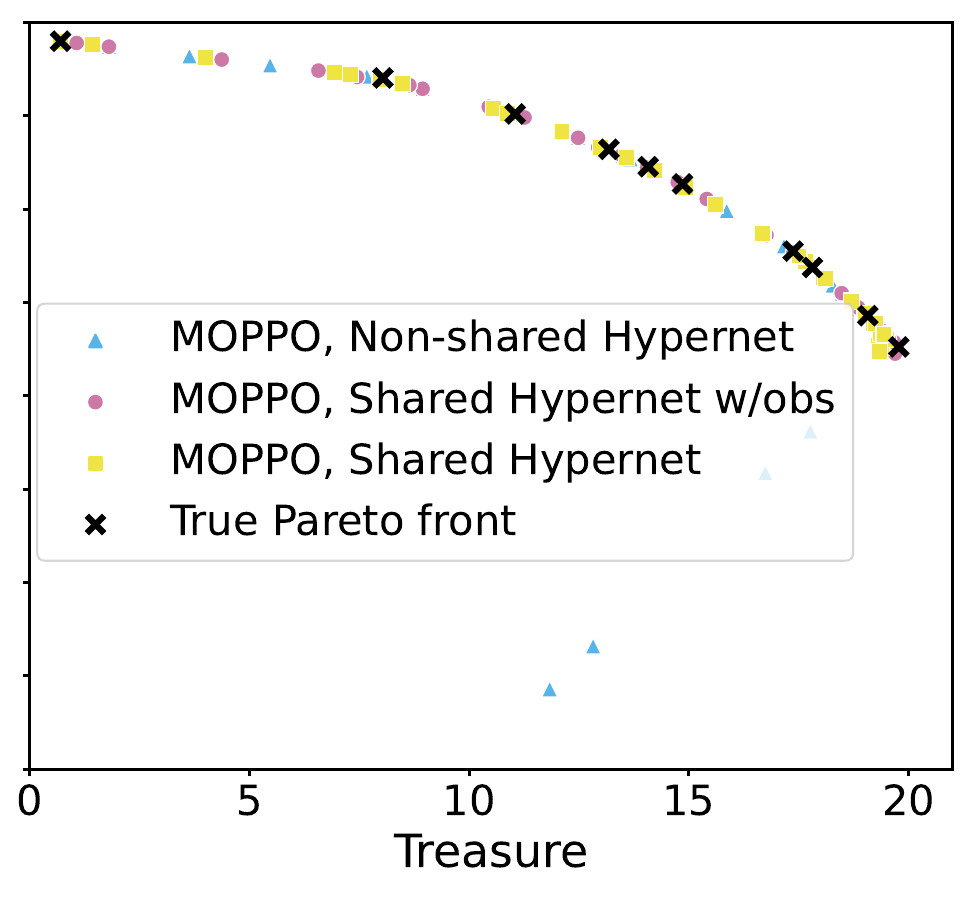}
      \caption{Hypernet architectures}
  \end{subfigure}

  \caption{\textbf{Pareto fronts on \textit{Deep Sea Treasure}:} Performance of a selection of our methods on the Deep Sea Treasure environment, split by the actor-critic architecture and the learning algorithm. The details of the architectures and algorithms are described in Sections~\ref{sec:arch} and~\ref{sec:algo}, respectively. In this simple gridworld, the agent's task is to find the biggest treasure, but big treasures require it to spend more fuel. The tension between the two objectives is formalized as a \emph{Pareto front} of the problem. Our proposed approaches effectively cover the true Pareto front. Some of the methods produce a few outliers because the policy struggles to learn near the boundary of the simplex $\Delta_K$ of reward weights, where one of the rewards (in this case, fuel) is completely discarded.}
  \label{fig:dst-results}
\end{figure}

Rather than maintaining a set of weights for each trade-off, we implicitly model the optimal set of policies by conditioning the learned policy on each objective's vector of \textit{relative weights}. We call this approach \emph{Dynamic MORL (DMORL)}. This allows us to learn a \textbf{single model} \emph{that encompasses all possible solutions on the Pareto front}. Our approach relies on \emph{linear scalarization} to model trade-offs. Thus, the policy learns to optimize a convex combination of the objectives and is conditioned on the coefficients of this combination. Despite the theoretical limitations of linear scalarization in MORL as noted by \cite{vamplew2008limitations}, we find that DMORL when paired with a sufficiently expressive neural network, can generate a continuous parameterization of the entire Pareto front. 
In this paper, we investigate two components needed to scale MORL for modern reinforcement learning tasks: the learning algorithm and the architecture. For the DMORL framework we explore in this paper, we mainly focus on PPO~\citep{schulman2017proximal} and generalize it to the multi-objective case. As a baseline, we introduce a multi-objective version of A2C~\citep{mnih2016asynchronous}.
We investigate multiple actor-critic architectures, including a multi-body network, merge networks for relative weights, and hypernetworks. We also normalize the rewards using the PopArt scheme~\citep{hessel2019multi} and propose a novel method to control the entropy during training. An example Pareto front produced by our methods on a simple test environment is shown in Figure~\ref{fig:dst-results}.
To summarize, our main contributions are:
\begin{itemize}
\itemsep 0em 
    \item We propose a scalable family of algorithms for multi-objective on-policy RL.
    \item We propose and evaluate different actor-critic architectures for multi-objective RL.
    \item We describe a method to control the policy's entropy during training for MORL dynamically and show that it improves learning stability.
    \item We demonstrate the effectiveness of our proposed approaches to DMORL on both deterministic and stochastic MORL benchmark environments.
\end{itemize}

\section{Related work}
\label{sec:relwork}

Multi-objective RL has been primarily studied from an off-policy perspective. Multiple MORL approaches have been developed as extensions of Q-learning~\citep{abels2019dynamic, lu2022multi}, some of which condition the Q-network on the scalarization weights as we do with policies. Among these methods, we use Envelope Q-learning~\citep{yang2019generalized} as a baseline, since its implementation is publicly available and includes the case of discrete actions, which is the focus of our work. Another notable example is the recent work by~\cite{hung2022q}, which also proposes a way to perform policy updates for policies conditioned on relative weights along with learning a Q-function, but the implementation only considers continuous control. Although Q-learning thrives in toy problems and is efficient in some more complex domains~\citep{mnih2013playing}, it has many failure modes, including the so-called ``deadly triad''~\citep{van2018deep}. Off-policy methods also struggle with capacity loss~\citep{lyle2022understanding} and, more generally, with generalizable feature learning~\citep{lan2022generalization}. Our work instead focuses on on-policy methods.

\cite{alegre2023sample} also propose a sample-efficient method for MORL, but they employ model-based learning while we focus on the model-free setup. \cite{roijers2018multi,reymond2023actor} directly optimize the policy in the model-free setting (the latter work also includes a critic), but target a single non-linear utility function. \cite{xu2020prediction}, similarly to us, apply PPO to the multi-objective case, but they do not condition the policies on the utility function, instead maintaining a set of policies explicitly. MOMPO by~\cite{abdolmaleki2020distributional} is an approach to MORL using techniques from distributional RL~\citep{levine2018reinforcement}. In contrast to our method, MOMPO requires multiple training runs to cover the Pareto front. 
One of the most relevant approaches to ours is called \emph{Pareto Conditioned Networks (PCN)} by~\cite{reymond2022pareto}. It operates in the DMORL setup and learns a policy conditioned on the desired value for all objectives. This allows for PCN to generalize to non-convex PF. This work draws from reward-conditioned policies (RCP) by~\cite{kumar2019reward}, a recent supervised learning algorithm for RL. However, due to the nature of its replay buffer, PCN is not adapted to stochastic environments.

\section{Dynamic multi-objective reinforcement learning}
\label{sec:dmorl}
Multi-objective reinforcement learning is the search for optimal policies for multi-objective Markov decision processes (MOMDP). A MOMDP is a tuple $\left( \ssp,\A,\rw, P, \kappa,\gamma \right) $, where sets  $\ssp$ and $\A$ are state and action spaces respectively, $\rw: \ssp\times\A\to\R^K$ is a $K$-dimensional reward function, and $P(s'\mid s,a)$ is the transition probability. Finally, $\kappa$ is the distribution of initial states, and $\gamma\in[0,1)$ is the discount factor. The only difference between MOMDP and MDP is the range of the reward --- $\R^K$ instead of $\R$. In this work, we optimize so-called linear scalarizations with relative weights given by $\balpha\in\dk$ from the $(K-1)$-dimensional simplex
\begin{equation}
  \dk=\left\{ \balpha \in\R^K \mid \sum_{i=1}^{K} \alpha_i=1, \ \alpha_i\ge 0\ \forall i \right\} .
\end{equation}
For $\balpha\in\dk$, a \emph{scalarized reward} is $r(s,a,\balpha)=\balpha^{\top} \mathbf{r}(s,a)$. Since we would like to cover all scalarizations with a single model, we have to generalize our policy definition so that it is also conditioned on $\balpha$. Hence, our parameterized policies are of the form $\pi(a\mid s, \balpha)$. For a policy $\pi$, the vector-value function also has to depend on $\balpha$:
\begin{equation}
  \vpi(s,\balpha)=\E_{\tau\sim p^\pi(\tau\mid \balpha)}\left[ \left.\sum_{t=0}^{\infty } \gamma^t \rw(s_t, a_t)\right| s_0=s \right] \in R^K.
\end{equation}
Here, $\tau=(s_0,a_0,s_1,\ldots)$ is a trajectory sampled from the transition dynamics $P$ and the policy $\pi(a\mid s,\balpha)$. We can also define the Q-function and the advantage in the usual way:
\begin{gather}
  \qpi(s,a,\balpha)=\rw(s,a)+\gamma\E_{s'\sim P(s'|s,a)}\left[\vpi(s',\balpha)\right], \\
  \api(s,a,\balpha)=\qpi(s,a,\balpha)-\vpi(s,\balpha).
\end{gather}
The vector-objective associated with $\pi$ is then given by 
 \begin{equation}
   \J(\pi,\balpha)=\E_{s\sim\kappa}\left[ \vpi(s,\balpha) \right] \in\R^K.
\end{equation}
In practice, our policies will be represented by a neural architecture with parameters $\theta\in\R^{D_a}$. In this case, we write the policy, a.k.a. the \emph{actor}, as $\pi_\theta(a|s,\balpha)$ and use shortcuts  $\vt=\mathbf{V}^{\pi_\theta}$ and $\mathbf{J}(\theta,\balpha)=\mathbf{J}(\pi_\theta, \balpha)$. MOPPO and MOA2C will also require a \emph{critic}, i.e., a neural approximation to the value function. It is parameterized by $\psi\in\R^{D_c}$, and we denote it as $\vc(s, \balpha)$. Some of the parameters might be shared between the actor and the critic. Our architecture's overall set of parameters will be called $\nu$. 

\paragraph{Pareto Front} To define a Pareto Front, we need the notion of strict dominance. We say that a vector $\p\in\R^K$ \emph{strictly dominates} $\q\in\R^K$ (denoted as $\p\succ \q$) iff $p_i>q_i$ for all $i\in\{1,\cdots K\}$. Next, individual parameters $\balpha$ give rise to partially specified policies $\pi(\cdot\mid\cdot,\balpha)$. We say that a policy $\pi(\cdot\mid\cdot,\balpha)$ strictly dominates $\pi(\cdot\mid\cdot,\balpha')$ iff $\E_{s\sim\kappa}\big[\mathbf{V}^{\pi}(s,\balpha)\big] \succ \E_{s\sim\kappa}\big[\mathbf{V}^{\pi}(s,\balpha')\big]$. A partially specified policy $\pi(\cdot\mid\cdot,\balpha)$ is said to be \emph{weakly Pareto-optimal} if no other policy dominates it. The set of all such policies is called a \emph{weak Pareto set}, and the image of the Pareto set in the objective space is called a \emph{weak Pareto front} (PF).

In a nutshell, to identify the points on the PF, we are optimizing the \emph{expected utility metric}, introduced by~\cite{zintgraf2015quality} as
\begin{equation}
  J(\theta)=\E_{\balpha\sim\D(\balpha)}\left[\balpha^{\top}\mathbf{J}(\theta,\balpha) \right]\label{eq:eum}
\end{equation}

for some distribution $\D$ over  $\dk$ with full support. Currently, we always set  $\D$ to be uniform, but other distributions can also be relevant. For a convex PF, if the neural architecture is expressive enough so that the entire PF is given by a single set of weights $\theta^*$, it is easy to see that $\theta^*$ will also be the maximizer of~\eqref{eq:eum}. The theoretical situation is more involved in the misspecified case (when the optimal policies are not represented by any $\theta$) since the maximizer depends on the exact distribution $\D$. Still, we will not study our approach from the theoretical perspective.

We evaluate our models with two of the most common metrics in the MORL literature: expected utility~\eqref{eq:eum} and hypervolume.
\paragraph{Hypervolume}
Given a Pareto front $\mathcal{P}$ and a reference point $\p_0$, the hypervolume $HV(\mathcal{P})$ indicates the volume of the polytope in the region of the objective space dominated by $\mathcal{P}$ and bounded from below by the reference point. Formally, in $M-$dimensional objective space, it can be defined as:
\begin{equation}
    HV(\mathcal{P}) = \int_{\mathbb{R}^M} \mathbb{I}_{H(\mathcal{P})}(\z)d\z
\end{equation}
where $H(\mathcal{P}) = \{\z \in \mathcal{Z}~|~\exists \p\in\mathcal{P}: \p_0 \preceq	\z \preceq \p \}$. Here, $\preceq$ is the relation operator that indicates the Pareto-dominance, and $\mathbb{I}_{H(\mathcal{P})}$ is the indicator function that is $1$ if $\z \in H(\mathcal{P})$, and $0$ otherwise.

\section{Architectures}
\label{sec:arch}

\definecolor{statecol}{HTML}{2E2585}
\definecolor{embeddingcol}{HTML}{434343}
\definecolor{alphacol}{HTML}{337538}
\definecolor{alphashadingcol}{HTML}{93C47D}
\definecolor{hypercol}{HTML}{94CBEC}
\definecolor{actorcol}{HTML}{7E2954}
\definecolor{criticcol}{HTML}{DCCD7D}

% \tcbset{
%     myhatchbox/.style={
%         enhanced,
%         boxrule=0pt,
%         colframe=white,
%         colback=white,
%         frame hidden,
%         interior style={pattern=grid, pattern color=gray!20},
%         sharp corners,
%         width=\linewidth,
%         boxsep=0pt,
%         left=0pt,
%         right=0pt,
%         top=0pt,
%         bottom=0pt
%     }
% }
% \newtcolorbox{mybox}{colback=red!5!white,
% colframe=red!75!black}
% \newtcbox{\mybox}[1][red]{on line,
% arc=0pt,outer arc=0pt,colback=#1!10!white,colframe=#1!50!black,
% boxsep=0pt,left=1pt,right=1pt,top=2pt,bottom=2pt,
% boxrule=0pt,bottomrule=1pt,toprule=1pt}
\newtcbox{\alphabox}[1][green]{
    on line,
    arc=3.5pt,
    colback=alphashadingcol,
    colframe=alphacol,
    boxrule=0.7pt,
    boxsep=0pt,
    left=2pt,
    right=2pt,
    top=1.5pt,
    bottom=1.5pt
}
\newcommand{\herealpha}{\alphabox{$\balpha$}}
% Alt
% \newcommand{\herealpha}{{\color{alphacol}$\balpha$}}

\newtcbox{\statebox}[1][purple]{
    on line,
    arc=0pt,
    colback=white,
    % opacityback=0.5,
    colframe=statecol,
    boxrule=0.7pt,
    boxsep=0pt,
    % interior style={pattern=vertical lines},
    left=2pt,
    right=2pt,
    top=1.5pt,
    bottom=1.5pt
}
\newcommand{\thestate}{\statebox{State $s_t$}}

\begin{figure}[t]
    \centering
    \begin{subfigure}[t]{0.33\linewidth}
        \centering
        \includegraphics[height=0.10\textheight]{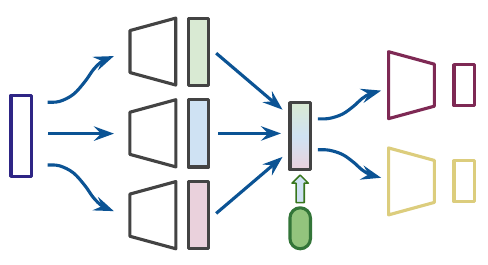}
        \caption{\textbf{Multi-body}. \hspace{-0.2em}\thestate\  is passed through individual embeddings, outputs are interpolated using \herealpha. Then, the interpolated outputs are passed to {the \color{actorcol}actor} and {\color{criticcol}critic} heads.}
    \end{subfigure}
    \hspace{1em}
    \begin{subfigure}[t]{0.28\linewidth}
        \centering
        \includegraphics[height=0.10\textheight]{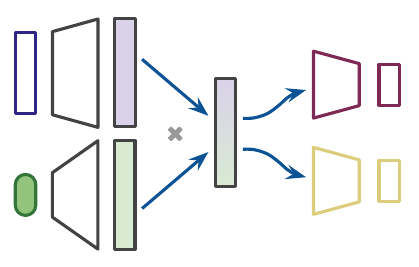}
        \caption{\textbf{Merge net}. \hspace{-0.15em}\thestate\  and \herealpha\  are passed through embeddings and multiplied. This is then passed to {the \color{actorcol}actor} and {\color{criticcol}critic}.}
    \end{subfigure}
    \hspace{1em}
    \begin{subfigure}[t]{0.28\linewidth}
        \centering
        \includegraphics[height=0.10\textheight]{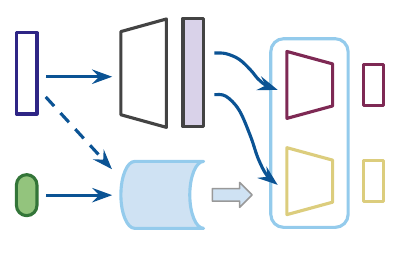}
        \caption{\textbf{Hypernetwork}. \thestate\  is embedded, and \herealpha\  is passed through a {\color{hypercol}hypernetwork} producing {the \color{actorcol}actor} and 
 {\color{criticcol}critic} heads.}
    \end{subfigure}
    \caption{\textbf{Actor-critic architectures with shared trunks:} Non-shared versions are organized similarly. The dashed line in the hypernetwork chart is optional: $s_t$ can be passed into the hypernetwork, where we get the architecture's ``hypernet w/obs'' variant.} \label{fig:arch}
\end{figure}

This work focuses on continuous state spaces and discrete actions, such that $\ssp\subset\R^{d_s},\quad\A=\left\{ 1,\ldots, K \right\}.$ Generalization to small discrete state spaces is straightforward with one-hot encoding, and generalization to continuous actions can be performed analogously to scalar PPO and A2C variants through parameterized action distributions. We consider three architectures for the actor-critic network, presented in Figure~\ref{fig:arch}.

When both actor and critic are learned, an often-used practice is to share weights between them. A popular approach is to share the \textit{trunk} $\bod: \ssp\to \R^F$ of a neural network that takes the state $s$ in and produces intermediate \emph{features} $\bod(s)$. The trunk is parameterized by $\zeta \in \R^{D_b}$. We hope the features are general enough to convey information about both the optimal action and the value of the current state. In this case, we can use a \emph{shared trunk} architecture with two separate linear layers to extract both the action distribution and the predicted value:
\begin{align}
  \pi_{\theta}(\cdot\mid s,\balpha)&=\text{softmax}(\mathbf{W}_a\bod(s,\balpha)+\mathbf{b}_a), \label{eq:ac} \\
  \vc(s,\balpha)&=\mathbf{W}_c\bod(s,\balpha)+\mathbf{b}_c. \label{eq:crit}
\end{align}
The learnable parameters of the actor are therefore $\theta=\left( \zeta,\mathbf{W}_a,\mathbf{b}_a \right) $. For the critic, the parameters are $\psi=\left( \zeta, \mathbf{W}_c,\mathbf{b}_c \right) $. We refer to $(\mathbf{W}_a,\mathbf{b}_a)$ as the \emph{actor head} and to $(\mathbf{W}_c,\mathbf{b}_c)$ as the \emph{critic head}. An alternative that we consider is that the critic and the actor do not share the trunk's parameters, even though its architecture is the same for both. We will refer to this as a \emph{non-shared trunk} architecture.

When the policy is conditioned on states $s\in\ssp$ and relative weights $\balpha\in\dk$, it is a priori unclear what woul  d be the best architecture that combines these inputs in $\bod$. Deep learning architectures provide a useful inductive bias to guide learning and cover the Pareto front efficiently. This can be especially helpful when the rewards interfere with each other, for example, when they compete, i.e., maximizing one reward minimizes the other\footnote{In Minecart environment, rewards for fuel and ores can have competing dynamics.}. As a result, the signs of each objective gradient maximizing each reward may differ, resulting in canceling each other in the shared trunk's parameters. This can disrupt the learning dynamics and representations. An effective way to address this in the continual multi-task learning has been to modify the architectures (see \citet{rusu2016progressive}.) Here, we consider three architectures to understand their impact on learning policies for DMORL: a \emph{multi-body network}, a \emph{merge network}, and a \emph{hypernetwork}.

\textbf{Multi-body network. }This architecture follows the intuition that the policies using different weights are, in some sense, interpolated. However, our preliminary experiments have shown that if we use multiple heads for actors corresponding to different objectives, the interpolation between logits or logprobs is not expressive enough. Instead, we need to interpolate in the higher-dimensional feature space. We thus settled on the formulation
\begin{equation}
  \bod(s,\w)=\operatorname{MLP}\left(\sum_{i=1}^{K} \alpha_i \phi\left( \mathbf{W}_{i}s+\mathbf{b}_{i} \right)\right).
\end{equation}
Here, $\phi$ is the ReLU activation~\citep{nair2010rectified}, $\balpha=(\alpha_1,\ldots, \alpha_k)^\tp$, and $\operatorname{MLP}$ is a multi-layer perceptron. This architecture has a ``body'' $(\mathbf{W}_{i},\mathbf{b}_i)$ for each objective, which independently processes the input state, and their outputs after activation are linearly interpolated by the weights $\balpha$ assigned for each objective.

\textbf{Merge network. } Here, similarly to~\cite{reymond2022pareto}, we pass $\balpha$ through a separate embedding before merging it with the state:
\begin{equation}
  \bod(s,\w)=\operatorname{MLP}\left(\tilde{\phi}(\mathbf{W}_1s+\mathbf{b}_1)\odot\tilde{\phi}(\mathbf{W}'_1\balpha+\mathbf{b}'_1)\right),
\end{equation}
where $\tilde{\phi}$ is the sigmoid activation, and $\odot$ stands for element-wise multiplication.

\textbf{Hypernetwork. }Finally, we consider a slightly more involved architecture similar to the one described in~\cite{navon2020learning}. Below, we detail the shared trunk case. The body of the policy $\bod(s)$ here only takes the state as input:
\begin{equation}
\bod(s)=\operatorname{MLP}(s).
\end{equation}
The last layer parameters  $\mathbf{W}_a, \mathbf{b}_a,\mathbf{W}_c,\mathbf{b}_c$ are now produced by a hypernetwork with the structure
\begin{gather}
  \mathbf{h}(\balpha)=\phi(\mathbf{W}_h\balpha+\mathbf{b}_h), \\
  \mathbf{W}_a,\mathbf{b}_a=\mathbf{W}_2\mathbf{h}+\mathbf{b}_2,\quad \mathbf{W}_c,\mathbf{b}_c=\mathbf{W}'_2\mathbf{h}+\mathbf{b}'_2.
\end{gather}
% \caglar{We should cite the multi-objective hypernetwork paper.} This is the Navon paper
Above, we slightly abuse the notation in the sense that $\mathbf{W}_3\mathbf{h}+\mathbf{b}_3$ is a vector of dimension $(F+1)|\A|$, which gets split and reshaped into the corresponding layer parameters $\mathbf{W}_a$ and $\mathbf{b}_a$.
% To avoid the explosion of the number of parameters, the used feature dimension $F$ is much smaller in this architecture than in the other architectures ($64$ instead of $256$). 
Note how the actor and the critic share the network's trunk and the first layer of the hypernetwork. With a non-shared trunk, the critic gets its copy of all parameter groups instead. We also experimented with a version of the shared trunk architecture where the observation gets fed into the hypernetwork along with the weights, i.e., $\mathbf{h}(\balpha,s)=\phi(\mathbf{W}_h[\balpha;s]+\mathbf{b}_h)$, abbreviated as ``Hypernetwork w/obs'' in the paper.

\section{Algorithms}
\label{sec:algo}

We chose to focus on two instantiations of our framework for DMORL. However, our actor-critic architectures are more general and could be applied to other policy iteration or policy gradient methods.
\subsection{Actor and critic losses}
\paragraph{Multi-objective policy gradient} Scalar A2C relies on the Policy Gradient theorem to update the actor. This theorem can be generalized to the vector-valued reward case:
\begin{align}
  \nabla_\theta (\balpha^{\top}\mathbf{J}(\theta, \balpha))&= \E_{\tau\sim p^\pi(\tau|\balpha)}\left[ \sum_{t=0}^{\infty } \gamma^t \balpha^\top \mathbf{A}^\theta(s_t,a_t,\balpha)\nabla_\theta \log\pi_\theta(a_t\mid s_t,\balpha)\right].
\end{align}
To maximize the expected utility, we approximate the above expression by first sampling random relative weight vectors $\balpha \sim \D(\dk)$, and then performing a rollout of the current policy $\pi^\theta$ using $\balpha$. In practice, we maintain $B$ trajectories and corresponding $\balpha$-s simultaneously, but in this section we set $B=1$ for clarity. Let the discounted reward-to-go for the trajectory at timestep $t$ be $\hq_{t}$. If the trajectory was truncated before the terminal state, reward-to-go is bootstrapped. The estimator of the gradient of expected utility that we use in multi-objective A2C is given by
\begin{align}
  \nabla_\theta J(\theta)\approx\hnab_\theta J(\theta)= \sum_{t=0}^{T } \gamma^t \balpha^\tp(\hq_t-\vc(s_{t}))\nabla_\theta \log\pi_\theta(a_{t}\mid s_{t},\balpha). \label{eq:a2c_est}
\end{align}
\paragraph{Critic loss} To optimize the critic over a minibatch of $\{(s_k,\hq_k)\}_k$, we use the least-squares loss:
\begin{equation}
  l_c(\psi)=\sum_{k}\norm{\hq_{k}-\vc(s_{k},\balpha)}^2. \label{eq:critic_loss}
\end{equation}

\begin{wrapfigure}[28]{R}{0.6\textwidth}
\hspace{1em}
\begin{algorithm}[H]
\DontPrintSemicolon
\SetNoFillComment
% \SetNlSty{}{}{}
\SetKwInput{KwRequire}{Require}
\KwRequire{Multi-objective MDP $\mathcal{M}=\left( \ssp,\A,\rw, P, \mu,\gamma \right)$}
\KwRequire{Actor $\pi_\theta$, critic $\vc$ with parameters $\nu=\theta\cup\psi$}
% Initialize $\nu$\;
\While{not terminated}{
    Sample reward weights $\balpha\sim U(\Delta_K)$\;
    Sample a truncated trajectory $\{(s_{t},a_{t},\rw_{t})\}_{t=0}^T$ \\
    \quad from $P$ and $\pi_\theta(\cdot\mid \cdot,\balpha)$\;
    \For(\tcp*[h]{iterate over epochs}){$e\gets 1$ \KwTo $E$ }{ 
        Update PopArt $\bmu$ and $\bsigma$ and the critic head  \;
        Get the reward-to-go $\hq_t$ from bootstrapped $\operatorname{TD}(1)$ \;
        Compute vector- and scalarized advantage: \\
        \quad $\hat{\mathbf{A}}_t\gets \hq_t-\vc(s_t,\balpha),\ \hA_t\gets\balpha^\tp(\hat{\mathbf{A}}_t/\bsigma)$ \;
        \For{$b\gets 1$ \KwTo $B$}{
            Sample a minibatch $\{(s_k,a_k,\hq_k,\hA_k)\}_k$\;
            Update $\g$ and $\lambda$ using~\eqref{eq:entropy_update} for entropy control\;
            Compute $l_a(\psi)$ and its gradient $\nabla_\theta l_a$ from~\eqref{eq:actor_loss}\;
            Compute $l_c(\psi)$ and its gradient $\nabla_\psi l_c$ from~\eqref{eq:critic_loss}\;
            Update $\beta_c$ so that $\norm{\g+\nabla_\theta l_a}\approx C\beta_c\norm{\nabla_\psi l_c}$\;
            Feed $\g+\nabla_\theta l_a+\beta_c\nabla_\psi l_c$ into Adam, update $\nu$\;
        }
    }
}
% \Return{$\nu$}\;
% \vspace{0.5em}
\caption{A schematic description of multi-objective PPO for an actor and a critic with a shared trunk. Our PopArt update scheme follows~\cite{hessel2019multi}. Actor-to-critic gradient ratio $C$ is one of the hyperparameters we tune.}
\label{alg:moppo}
\end{algorithm}
\end{wrapfigure}

The algorithm dynamically adjusts the weight $\beta_c$ of the critic loss. Depending on whether the critic shares parameters with the actor, it makes sense to use different strategies for setting $\beta_c$. For the generality of our methods, we set $\beta_c$ dynamically to ensure an approximately constant ratio of norms of the actor and critic gradients. This technique was not as important in our environments, so we defer a discussion to Appendix~\ref{app:combining}. 

\paragraph{PopArt} is an approach to learning value functions across different orders of magnitude, especially in situations where the value scales are not known in advance or change depending on the performance of the policy. \cite{hessel2019multi} successfully adapted it to the multi-task RL setup, similar to MORL. In MORL, the differences in scale between rewards also pose a problem. If we have to design trade-offs between objectives of varying magnitude, then the optimization target~\eqref{eq:eum} will unfairly favor the objectives of larger scales. Multi-task PopArt maintains an approximate mean $\mu_i$ and variance $\sigma_i^2$ of the target values for each task. 
We chose to use $\balpha$ combine the normalized advantages. This means that for the gradient estimate~\eqref{eq:a2c_est} instead of $\balpha^\tp(\hq_{t,j}-\vc(s_{t,j}))$ we use a scalarization of the form
\begin{equation}
  \balpha^\tp\left((\hq_{t,j}-\vc(s_{t,j}))/\bsigma\right), \label{eq:PopArt_scalarization}
\end{equation}
where $\bsigma=(\sigma_1,\ldots,\sigma_K)$ and $/$ denotes component-wise division.
\paragraph{Multi-objective PPO}
The original PPO is formulated by~\cite{schulman2017proximal} using a~\emph{surrogate objective} for policy iteration. This objective is optimized for multiple epochs over the same sampled trajectories to achieve higher sample efficiency. The actor loss for MOPPO differs from the standard PPO formulation only by using the scalarized advantage $\hA$, defined in Algorithm~\ref{alg:moppo}. Given a minibatch of $\{(s_k,a_k,\hA_k)\}_k$, we employ the following loss of the actor:
\begin{equation}
    l_a(\theta)=\sum_{k}\min\left(r_k(\theta)\hA_k,\operatorname{clip}\left(r_k(\theta), 1-\epsilon,1+\epsilon\right)\hA_k\right),\quad r_k(\theta)=\frac{\pi_{\theta}(a_k|s_k,\balpha)}{\pi_{ref}(a_k|s_k,\balpha)},\label{eq:actor_loss}
\end{equation}
where $\pi_{ref}$ is the policy directly after sampling the trajectory.

\subsection{Entropy control during training}
\label{sec:entropy}
Our experiments show that entropy regularization is necessary to avoid collapse in more challenging environments. The entropy of the current policy embodies the exploration-exploitation trade-off: if the entropy is low, the agent fails to explore, but if it is too high, the agent cannot operate efficiently. Intuitively, a good training run keeps entropy high initially to explore sufficiently and then ``anneals'' the entropy to lower values as training progresses. This behavior is hard to achieve in practice since training is sensitive to the entropy regularization coefficient. At this point, we can employ a key insight that we do not need to \emph{maximize} the entropy. Rather, we would like to keep it at a pre-defined level. Hence, the framework of constrained optimization is applicable. 

\begin{figure}[t]
  \centering
  \includegraphics[width=0.3\linewidth]{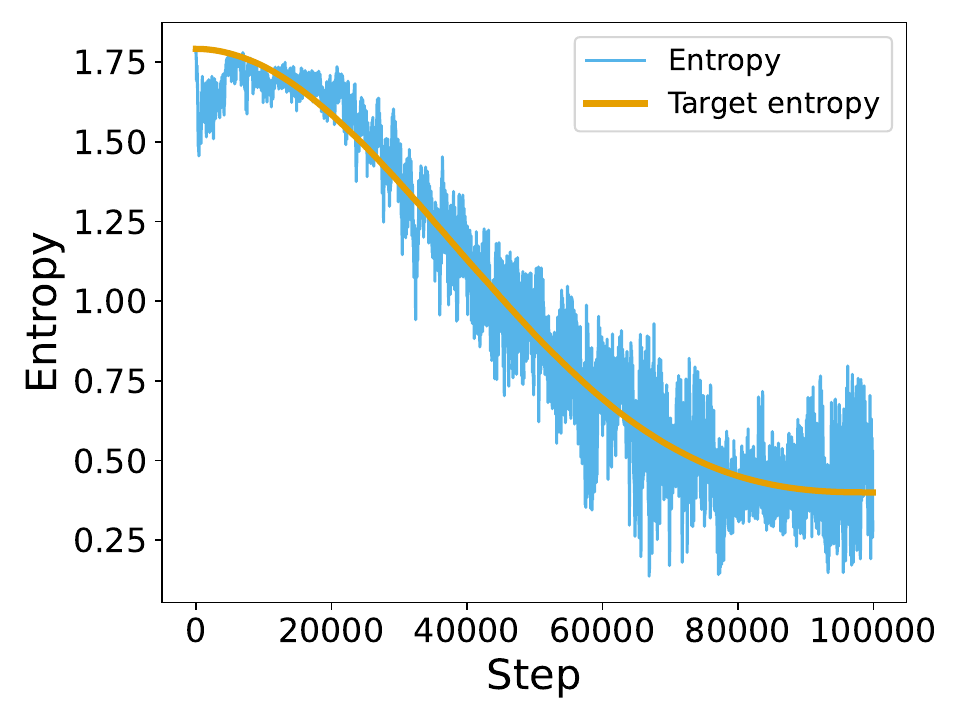}
  \includegraphics[width=0.3\linewidth]{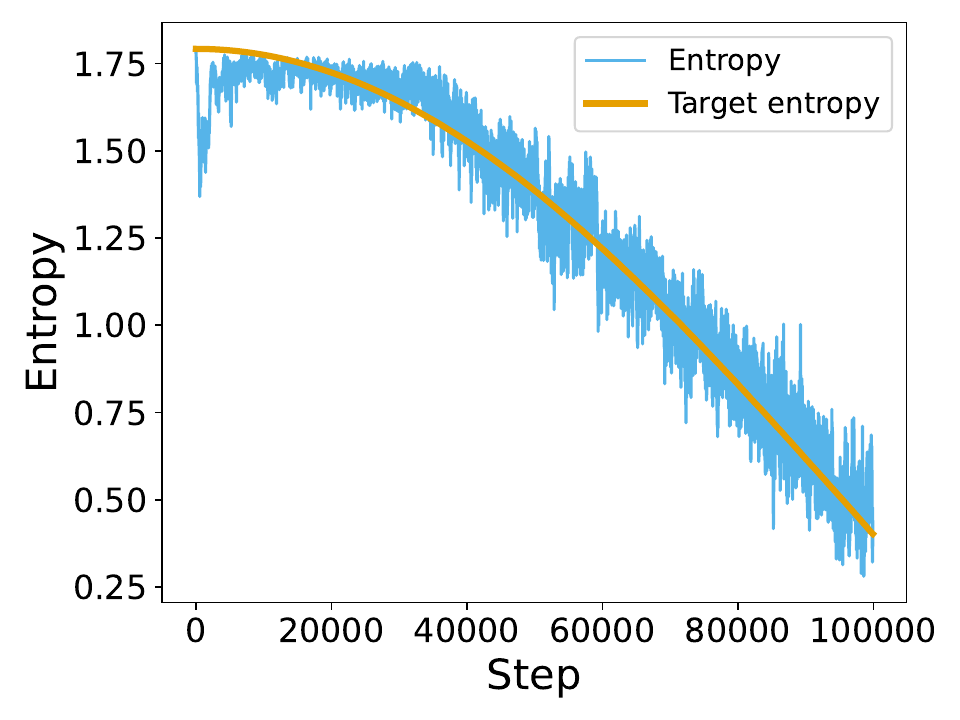}
  \includegraphics[width=0.3\linewidth]{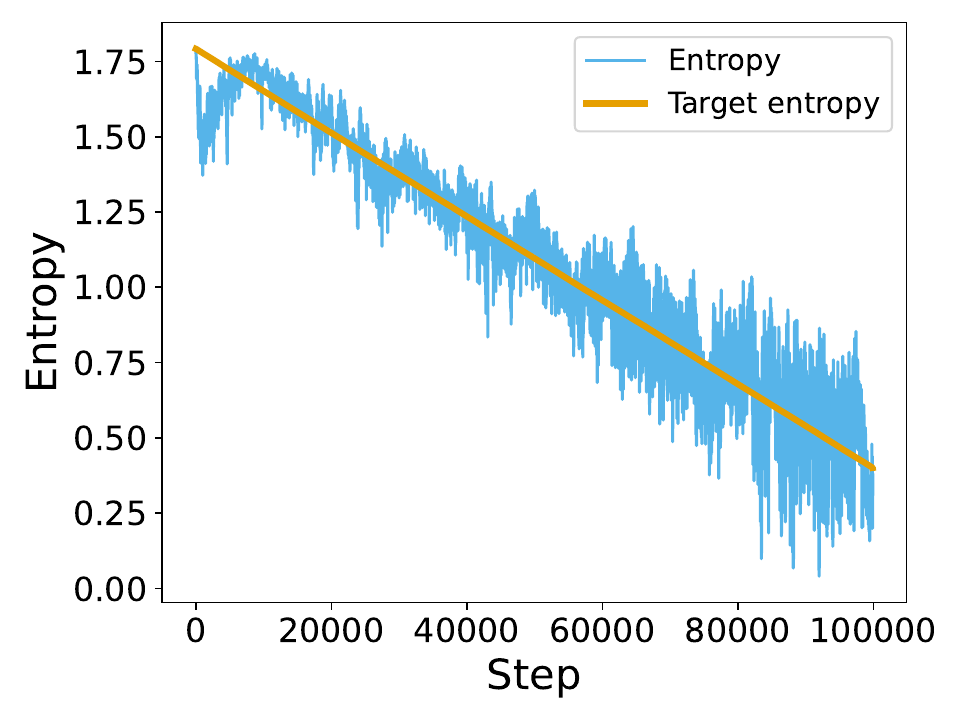}
  \caption{\textbf{Entropy control schedules:} Example entropy behaviors on Minecart when using the entropy control method described in Section~\ref{sec:entropy}. From left to right: custom schedule, cosine schedule and linear schedule of entropy. The custom schedule is designed to have a flat start for exploration and an extended flat end for fine-tuning the behavior. The schedules are discussed in detail in Appendix~\ref{app:entropy}.}
  \label{fig:entropy}
\end{figure}

We use an algorithm by~\cite{platt1987constrained} called the \emph{Modified Differential Method of Multipliers (MDMM)}, which allows us to maximize a function subject to approximate equality constraints. Let the expected entropy of the policy and its empirical estimate be 
\begin{equation}
    H(\theta)=\E_{\balpha\sim \D(\Delta_K),s\sim p^{\pi_\theta}(s|\balpha)}\left[H(\pi_\theta(\cdot\mid s,\balpha))\right],\quad \widehat{H}(\theta)=\frac{1}{T+1}\sum_{t=0}^{T}H(\pi_\theta(\cdot\mid s_{t},\balpha)),
\end{equation}
where $p^{\pi_\theta}$ is the state visitation distribution. Assume that we would like to satisfy the constraint $H(\theta)\approx H_{target}$, where $H_{target}$ is a desired entropy value that can also depend on our progress in training. To this end, MDMM introduces the Lagrange multiplier $\lambda_i\in\R$, which is dynamically updated at each step $i$. Vanilla MDMM dictates that we use an update of the form: 
\begin{equation}
    \g=(\lambda_i+c(H_{target}-\widehat{H}))\nabla_\theta\widehat{H},\quad
    \theta_{i+1}=\theta_{i}+\eta(\g+\nabla_\theta l_a),\quad \lambda_{i+1}=\lambda_i+\tilde{\eta}(H_{target}-\widehat{H}). \label{eq:entropy_update}
\end{equation}
Here, $\eta$ is the learning rate, and $\tilde{\eta},c$ are new hyperparameters. In practice, we don't directly use the update vector $\g+\nabla_\theta l_a$ and instead provide it to Adam. Note that the conventional method of entropy regularization keeps $\lambda$ constant and positive, while here it can also become negative, thus forcing the entropy to decrease to reach $H_{target}$. Figure~\ref{fig:entropy} demonstrates how entropy oscillates around the target value during training in the Minecart environment. We discuss the entropy schedules we used in more detail in Appendix~\ref{app:entropy}. In our experiments, entropy control did not perform well with A2C, so we only enabled it for PPO, which provides benefits as our ablation study demonstrates. 

\section{Experiments}
\label{sec:experiments}
We implement all architectures described above using the \texttt{TorchRL} library~\citep{bou2023torchrl}. As a source of MOMDP environments, we use \texttt{MO-Gymnasium}~\citep{Alegre2022bnaic}, the standard testbed in MORL. We run an ablation study to compare various architectures and algorithm details against each other and perform a comparison against two baselines, Pareto Conditioned Networks~\citep{reymond2022pareto} and Envelope Q-learning~\citep{yang2019generalized}. We chose these two approaches to MORL because, to the best of our knowledge, these are the most recent model-free MORL methods that condition a single policy (or value function) to generate the entire Pareto front and that have a public implementation supporting the case of continuous observations and discrete actions. We use the implementation provided in MORL-baselines~\citep{felten_toolkit_2023} for both methods.

\subsection{Preliminary results on deep-sea-treasure}
\label{sec:prelim-dst}
All our methods can solve the standard grid-world environment called ``Deep Sea Treasure''. In this environment, the agent controls a submarine on a 2D grid. The two objectives are the cost of fuel ($-1$ for every step) and the reward of a treasure that the agent can discover in pre-defined locations. When the agent is willing to spend more fuel, it can find a bigger reward. Precise control over the placement of treasures and the rewards from each one allows us to shape the 2D PF. We use the version of Deep Sea Treasure with a convex PF. We discovered that the results in this simple environment are not sensitive to hyperparameter tuning. All runs used $\approx 10^5$ environment steps and converged to Pareto fronts presented in Figure~\ref{fig:dst-results}.

\subsection{Ablation studies on Minecart}
\label{sec:hyperparams}
\begin{figure}[t]
  \centering
  \includegraphics[height=0.35\linewidth]{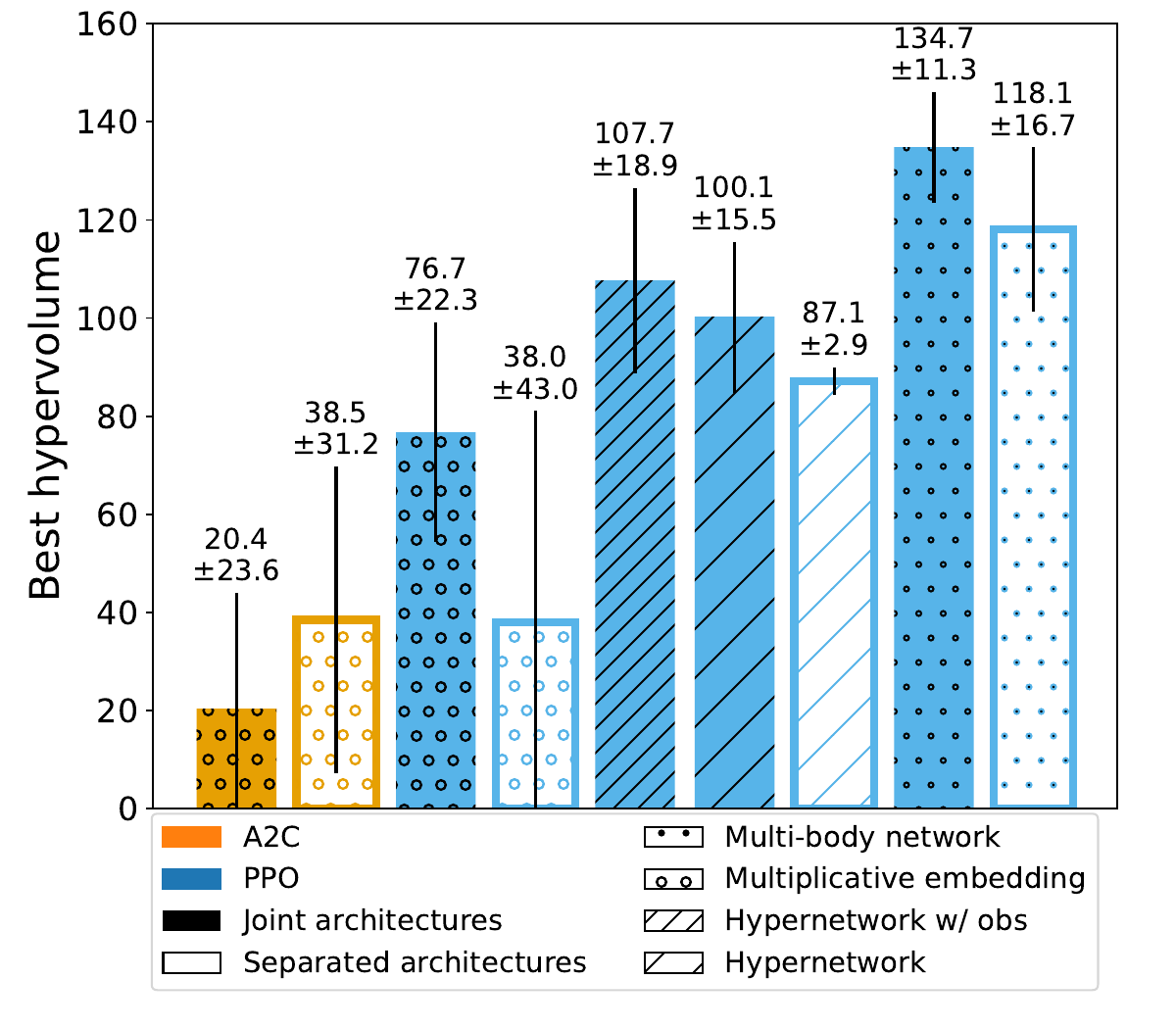}
  \hspace{1cm}
  \includegraphics[height=0.35\linewidth]{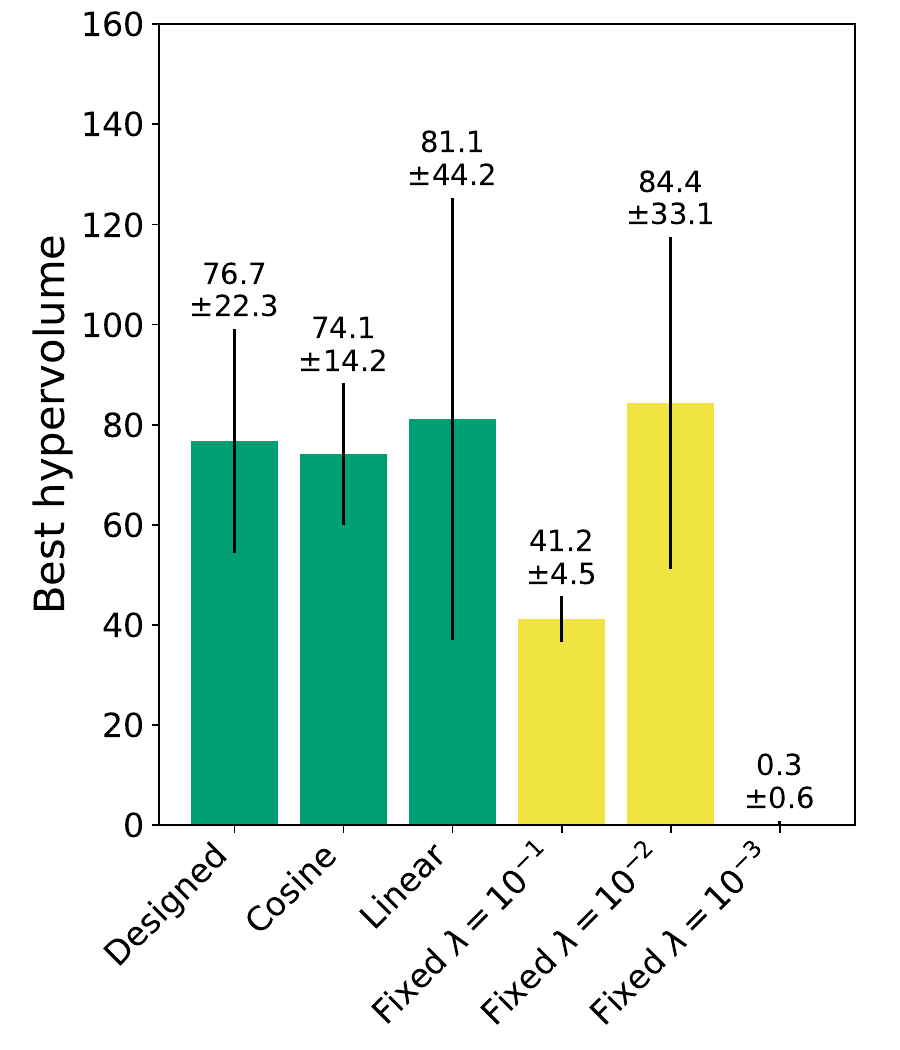}
  \caption{\textbf{Architecture ablations on Minecart:} \textit{Left:} comparison of the algorithms and policy architectures on Minecart. The color represents the algorithm (MOA2C or MOPPO), filled columns correspond to shared trunk architectures, and hatching denotes the specific architecture described in Section~\ref{sec:arch}. \textit{Right:} comparison of entropy control from Section~\ref{sec:entropy} using three entropy schedules described in Appendix~\ref{app:entropy} with standard entropy regularization using a fixed weight $\lambda$.}
  \label{fig:minecart-tuning}
\end{figure}

 ``Minecart'' is a harder environment than deep-sea-treasure, also included in MO-Gym. This environment has an agent that operates a cart in a 2D continuous space. There are three rewards: two for bringing different types of ores and one for the consumed fuel. Depending on the agent's location, the probabilities for getting one or the other change, thus leading to a nontrivial multi-objective problem. This environment also has a deterministic version, where each mining action near a mine is guaranteed to produce a fixed amount of ore. We use the stochastic version for ablations on different architectural choices discussed here. For each architecture, we run a hyperparameter search detailed in Appendix~\ref{app:hyper_arch}. Then, we run training for each method with five seeds to estimate the average and standard deviation. The resulting hypervolume for selecting methods is presented in Figure~\ref{fig:minecart-tuning}. We want to focus the attention of the reader on multiple conclusions that we can draw if we focus on parts of it: 
 % \caglar{Check if these are still true:}

\begin{itemize}
\itemsep 0pt
  \item PPO significantly outperforms A2C, likely because of its sample-efficient data reuse across multiple epochs. This effect mirrors the corresponding knowledge from the scalar RL community.
  \item Non-shared trunk architectures consistently perform slightly worse than their shared counterparts.
  \item Architecture choice seems to influence the performance. Multi-body networks perform the best, followed by hypernetworks and merge networks. This ordering also depends on the environment since it comes out differently for another environment in the following section.
\end{itemize}

Figure~\ref{fig:minecart-tuning} also shows that training is sensitive to the entropy control coefficient $\lambda$ when using standard entropy regularization. All 12 considered hyperparameter configurations lead to collapse if we fix $\lambda=10^{-3}$, while $\lambda=10^{-2}$ seems to lead to optimal performance among methods with constant $\lambda$. On the other hand, all entropy control schedules lead to essentially the same performance. We additionally note here that we did not have to perform tuning of the entropy control learning rate $\tilde{\eta}$ for any of our experiments, we just fixed $\tilde{\eta}=\eta/10$ from the start. We also provide the results in expected utility for all of the methods from Figure~\ref{fig:minecart-tuning} in Table~\ref{tab:big_table} in the Appendix \ref{app:normalization_ablation} and demonstrate all hyperparameter grids in Figures~\ref{fig:hyperparam_stoch_1} and~\ref{fig:hyperparam_stoch_2} there.

\subsection{Comparison with the baseline methods}
\label{sec:comp_baselines}
We use three environments for comparison: two versions of Minecart (deterministic and non-deterministic) and MO-reacher.  The non-deterministic version of Minecart checks that our methods can handle stochasticity, but it cannot be used to compare with PCN, which requires deterministic transitions. MO-reacher is a MuJoCo-based~\citep{todorov2012mujoco} environment where an agent controls a two-jointed robot arm. There are four rewards, each corresponding to the $l_2$-distance of the tip of the arm from one of four targets on the 2D-space. The action space consists of nine actions corresponding to one of three torques (positive, negative, and zero) in each joint.

 We tune the hyperparameters of each method on each environment separately; the details are presented in Appendix~\ref{app:hyper}. Results from all grids for this experiment are also presented in Figures 7-10 in the Appendix. The best hyperparameters are used to run the method again with $5$ different seeds. We found that in some cases, envelope Q-learning showed good results early but collapsed later in training, likely due to catastrophic forgetting. Because of this, we demonstrate the results corresponding to the best hypervolume for envelope Q-learning. Our methods or PCN did not suffer from this issue, so we reported the metrics after the training was finished. We present the evaluation results in Table~\ref{tab:baselines}. Overall, our methods can outperform the baselines in terms of hypervolume. This is partly because we use non-deterministic policies for evaluation, while PCN and Envelope Q-learning rely on deterministic ones. Learning non-deterministic policies allows us to cover the Pareto front more densely. As one can see, on Minecart, whether our methods outperform Envelope Q-learning is ambiguous. Mirroring the situation in the scalar RL literature, however, our methods perform better on the robotic control task. Note that the authors of PCN also provide hypervolume measurements on deterministic Minecart in~\cite[Table~1]{reymond2022pareto}, but they do not mention the reference point $\p_0$ they use, nor the discount factor $\gamma$. Based on the true Pareto front that they provide and their measurements, they most likely use $\gamma=1$, while we use $\gamma=0.99$. Therefore, the results from their paper are not directly comparable with the ones provided in our Table~\ref{tab:baselines}. 

\newcommand{\fst}[1]{\mathbf{#1}}
\newcommand{\snd}[1]{\underline{#1}}

\newcolumntype{C}{@{\hskip 0.25em}c@{\hskip 0.25em}}
\begin{table}[t]
    \centering
    \resizebox{\linewidth}{!}{
    \begin{tabular}{CCCCCCCCCCCCC}
    % \addtolength{\tabcolsep}{-0.3em}
\multicolumn{3}{c}{Method} &&& \multicolumn{2}{c}{Minecart} && \multicolumn{2}{c}{Minecart det.} && \multicolumn{2}{c}{Reacher} \\
\cmidrule{1-3} \cmidrule{6-7} \cmidrule{9-10} \cmidrule{12-13}
Arch & Shared & Entropy &&& HV & EU && HV & EU && HV ($/10^7$) & EU \\
\hline
\hline
\addlinespace[0.5em]
Multi-body & No & Custom &&& $\fst{115.8 \pm 15.5}$ & $0.19 \pm 0.06$ && $\fst{119.6 \pm 17.0}$ & $0.21 \pm 0.05$ && $\fst{3.21 \pm 0.64}$ & $\fst{21.06 \pm 3.23}$ \\
Multi-body & Yes & Custom &&& $\fst{132.7 \pm 11.3}$ & $0.27 \pm 0.02$ && $\fst{136.2 \pm 10.1}$ & $0.27 \pm 0.02$ && $2.64 \pm 0.87$ & $\fst{19.09 \pm 3.48}$ \\
Multi-body & Yes & Linear &&& $\fst{118.5 \pm 24.7}$ & $0.15 \pm 0.09$ && $\fst{134.6 \pm 8.8}$ & $0.27 \pm 0.02$ && $2.53 \pm 0.72$ & $\fst{18.26 \pm 3.64}$ \\
Hypernet & Yes & Custom &&& $97.9 \pm 16.5$ & $0.18 \pm 0.03$ && $93.8 \pm 6.0$ & $0.19 \pm 0.04$ && $2.45 \pm 0.91$ & $\fst{17.34 \pm 4.99}$ \\
Merge net & Yes & Custom &&& $84.2 \pm 14.1$ & $0.03 \pm 0.05$ && $50.4 \pm 42.1$ & $-0.22 \pm 0.26$ && $\fst{3.33 \pm 0.17}$ & $\fst{21.73 \pm 0.72}$ \\
% Merge net & Yes & Custom &&& $59.5 \pm 33.9$ & $0.08 \pm 0.09$ && $74.0 \pm 37.3$ & $0.12 \pm 0.12$ && $3.31 \pm 0.10$ & $21.67 \pm 0.44$ \\
\addlinespace[0.2em]
\hline
\addlinespace[0.2em]
\multicolumn{3}{c}{PCN} &&& --- & --- && $106.7 \pm 10.0$ & $\fst{0.30 \pm 0.06}$ && $1.91 \pm 0.23$ & $15.41 \pm 1.72$ \\
\multicolumn{3}{c}{Envelope} &&& $104.7 \pm 9.5$ & $\fst{0.34 \pm 0.03}$ && $77.4 \pm 4.7$ & $\fst{0.38 \pm 0.07}$ && $2.73 \pm 0.60$ & $\fst{19.78 \pm 2.65}$ \\
    \end{tabular}
    }
    \caption{A comprehensive comparison of a selection of our methods to baselines. We do not evaluate PCN in the nondeterministic minecart environment because the authors of the method claim that its handling of the replay buffer would lead to incorrect behaviors in stochastic environments. The best in the column is marked in bold, and so are all methods whose intervals intersect the interval of the best. 
    % \caglar{Also highlight the best numbers by making them bold perhaps.} \caglar{Explain in the caption why we don't have numbers of minecart for PCN.}
    }
    \label{tab:baselines}
\end{table}

\vspace{-0.5em}
\section{Conclusion}
\label{sec:conclusion}
We proposed several actor-critic architectures and two algorithms for dynamic multi-objective reinforcement learning. All of our approaches can solve a simple MORL environment and provide a continuous parametrization of the space of policies covering the Pareto front. We then performed an extensive comparison study on two more complicated environments, demonstrating the usefulness of the proposed improvements over the naive implementation. We showed that our implementation of multi-objective PPO can outperform the baselines on a robotic control task and perform competitively when delayed rewards and/or stochasticity are present in the environment.

\vspace{-0.5em}
\section{Limitations}
\label{sec:limitations}
Our contributions are empirical in nature. We compare architectures and technical details of the implementation, but we do not perform a theoretical analysis of our algorithms. Such analysis could also be beneficial for the community: for example, under which conditions does policy gradient or policy iteration converge to a policy that approximately covers the Pareto front, and what would the notion of ``misspecification'' be? Another important limitation of this work is that we only consider linear scalarizations of the vector return $\J(\pi,\balpha)$. This does not allow us to recover concave regions on the Pareto front. It should also be possible to extend our methods to nonlinear utilities, so that they optimize the expected scalarized return $\E\left[u\left(\balpha,\sum_t \gamma^t \rw_t\right)\right]$ for a utility $u$ with a free parameter $\balpha$. Finally, scalar PPO works especially well in continuous control environments, and a good extension of our work would be to see whether our methods preserve this strength in the multi-objective setup.

\bibliography{main}
\bibliographystyle{abbrvnat}

%%%%%%%%%%%%%%%%%%%%%%%%%%%%%%%%%%%%%%%%%%%%%%%%%%%%%%%%%%%%
\clearpage

\appendix
\section{Implementation details}
\subsection{Other variants of our learning algorithms}
Here we describe our versions of non-shared trunk PPO (Algorithm~\ref{alg:moppo_nonshared}) and of A2C (shared trunk version in Algorithm~\ref{alg:moa2c_shared}, and non-shared --- in Algorithm~\ref{alg:moa2c_nonshared}). We note that non-shared versions can perform multiple updates on the critic per single actor update, but this would be harder to justify conceptually for a shared trunk architecture.

Since MOA2C diverges when used together with our entropy regularization scheme, for it we have to rely on the standard entropy regularizer loss. Given a trajectory $\{(s_t,a_t,\rw_t)\}_t$ sampled with reward weights $\balpha$, it is given by
\begin{equation}
    \g_{a2c}=\frac{1}{T+1}\sum_{t=0}^T \nabla_\theta H(\pi_\theta(\cdot\mid s_t,\balpha). \label{eq:entropy_standard}
\end{equation}
\begin{algorithm}[t]
\DontPrintSemicolon
\SetNoFillComment
% \SetNlSty{}{}{}
\SetKwInput{KwRequire}{Require}
\KwRequire{Multi-objective MDP $\mathcal{M}=\left( \ssp,\A,\rw, P, \mu,\gamma \right)$}
\KwRequire{Actor $\pi_\theta$, critic $\vc$ with disjoint sets of parameters}
% Initialize $\nu$\;
\While{not terminated}{
    Sample reward weights $\balpha\sim U(\Delta_K)$\;
    Sample a truncated trajectory $\{(s_{t},a_{t},\rw_{t})\}_{t=0}^T$ \\
    \quad from $P$ and $\pi_\theta(\cdot\mid \cdot,\balpha)$\;
    \For(\tcp*[h]{iterate over epochs}){$e\gets 1$ \KwTo $E$ }{ 
        Update PopArt $\bmu$ and $\bsigma$ and the critic head  \;
        Get the reward-to-go $\hq_t$ from bootstrapped $\operatorname{TD}(1)$ \;
        Compute vector- and scalarized advantage: \\
        \quad $\hat{\mathbf{A}}_t\gets \hq_t-\vc(s_t,\balpha),\ \hA_t\gets\balpha^\tp(\hat{\mathbf{A}}_t/\bsigma)$ \;
        \For(\tcp*[h]{We run more updates for the critic}){$i\gets 1$ \KwTo $F$}{
            \For{$b\gets 1$ \KwTo $B$}{
                Sample a minibatch $\{(s_k,a_k,\hq_k)\}_k$\;
                Compute $l_c(\psi)$ and its gradient $\nabla_\psi l_c$ from~\eqref{eq:critic_loss}\;
                Feed $\nabla_\psi l_c$ into critic's Adam, update $\psi$\;
            }
        }
        \For{$b\gets 1$ \KwTo $B$}{
            Sample a minibatch $\{(s_k,a_k,\hA_k)\}_k$\;
            Update $\g$ and $\lambda$ using~\eqref{eq:entropy_update} for entropy control\;
            Compute $l_a(\psi)$ and its gradient $\nabla_\theta l_a$ from~\eqref{eq:actor_loss}\;
            Feed $\g+\nabla_\theta l_a$ into actor's Adam, update $\theta$\;
        }
    }
}
% \Return{$\nu$}\;
% \vspace{0.5em}
\caption{A schematic description of multi-objective PPO for an actor and a critic with separate trunks. In this case, the actor-to-critic ratio $C$ that we tune stands for the ratio $\eta_a/\eta_c$, where $\eta_a$ and $\eta_c$ are the learning rates of the actor and the critic optimizers, respectively. Since we tend to set $C>1$, we compensate here by running $F$ times more updates on the critic. In all our experiments we set $F=2$.}
\label{alg:moppo_nonshared}
\end{algorithm}

\begin{algorithm}[t]
\DontPrintSemicolon
\SetNoFillComment
% \SetNlSty{}{}{}
\SetKwInput{KwRequire}{Require}
\KwRequire{Multi-objective MDP $\mathcal{M}=\left( \ssp,\A,\rw, P, \mu,\gamma \right)$}
\KwRequire{Actor $\pi_\theta$, critic $\vc$ with disjoint sets of parameters}
% Initialize $\nu$\;
\While{not terminated}{
    Sample reward weights $\balpha\sim U(\Delta_K)$\;
    Sample a truncated trajectory $\{(s_{t},a_{t},\rw_{t})\}_{t=0}^T$ \\
    \quad from $P$ and $\pi_\theta(\cdot\mid \cdot,\balpha)$\;
    \For(\tcp*[h]{iterate over epochs}){$e\gets 1$ \KwTo $E$ }{ 
        Update PopArt $\bmu$ and $\bsigma$ and the critic head  \;
        Get the reward-to-go $\hq_t$ from bootstrapped $\operatorname{TD}(1)$ \;
        Compute vector- and scalarized advantage: \\
        \quad $\hat{\mathbf{A}}_t\gets \hq_t-\vc(s_t,\balpha),\ \hA_t\gets\balpha^\tp(\hat{\mathbf{A}}_t/\bsigma)$ \;
        Compute $l_c(\psi)$ and its gradient $\nabla_\psi l_c$ from~\eqref{eq:critic_loss} on the entire trajectory\;
        Compute the entropy regularizing gradient $\g_{a2c}$ from~\eqref{eq:entropy_standard} \;
        Compute the policy gradient estimator $\hnab_\theta J(\theta)$ from~\eqref{eq:a2c_est}\;
        Update $\beta_c$ so that $\Vert\g_{a2c}+\hnab_\theta J(\theta)\Vert\approx C\beta_c\norm{\nabla_\psi l_c}$\;
        Feed $\g_{a2c}+\hnab_\theta J(\theta)+\beta_c\nabla_\psi l_c$ into Adam, update $\nu$\;
    }
}
% \Return{$\nu$}\;
% \vspace{0.5em}
\caption{A schematic description of multi-objective A2C for an actor and a critic with shared trunks. We discovered that entropy control leads to divergence of MOA2C on our environments, so we disabled it in our experiments.}
\label{alg:moa2c_shared}
\end{algorithm}

\subsection{Gradients for shared and non-shared trunk architectures}
\label{app:combining}
Suppose the actor and critic do not share parameters. In that case, updating them in PPO or A2C is straightforward: the gradient of the value loss is separated from the gradient of the policy with entropy regularization. We can even perform multiple ``inner'' optimization steps on the critic. Since the critic is aiming at a moving target (the value function depends on the policy, which keeps changing), there should be a clear sweet spot between overfitting the critic to the current policy and making it unable to catch up. Although more extensive experiments would be helpful, for now, we just set the number of inner updates to $2$ (relative to one update of the actor) and only studied the relative learning rate we can give to the critic for optimal performance.

\begin{algorithm}[t]
\DontPrintSemicolon
\SetNoFillComment
\SetNlSty{}{}{}
\SetKwInput{KwRequire}{Require}
\KwRequire{Multi-objective MDP $\mathcal{M}=\left( \ssp,\A,\rw, P, \mu,\gamma \right)$}
\KwRequire{Actor $\pi_\theta$, critic $\vc$ with parameters $\nu=\theta\cup\psi$}
Initialize $\nu$\;
\While{not terminated}{
    Sample reward weights $\balpha\sim U(\Delta_K)$\;
    Sample a truncated trajectory $\{(s_{t},a_{t},\rw_{t})\}_{t=0}^T$ \\
    \quad from $P$ and $\pi_\theta(\cdot\mid \cdot,\balpha)$\;
    \For(\tcp*[h]{iterate over epochs}){$e\gets 1$ \KwTo $E$ }{ 
        Update PopArt $\bmu$ and $\bsigma$ and the critic head  \;
        Get the reward-to-go $\hq_t$ from bootstrapped $\operatorname{TD}(1)$ \;
        Compute vector- and scalarized advantage: \\
        \quad $\hat{\mathbf{A}}_t\gets \hq_t-\vc(s_t,\balpha),\ \hA_t\gets\balpha^\tp(\hat{\mathbf{A}}_t/\bsigma)$ \;
        \For{$i\gets 1$ \KwTo $F$}{
            Compute $l_c(\psi)$ and its gradient $\nabla_\psi l_c$ from~\eqref{eq:critic_loss} on the entire trajectory\;
            Feed $\nabla_\psi l_c$ into critic's Adam, update $\psi$\;
        }
        Compute the entropy regularizing gradient $\g_{a2c}$ from~\eqref{eq:entropy_standard} \;
        Compute the policy gradient estimator $\hnab_\theta J(\theta)$ from~\eqref{eq:a2c_est}\;
        Feed $\g_{a2c}+\hnab_\theta J(\theta)$ into actor's Adam, update $\theta$\;
    }
}
% \Return{$\nu$}\;
% \vspace{0.5em}
\caption{A schematic description of multi-objective A2C for an actor and a critic with non-shared trunks. We discovered that entropy control leads to divergence of MOA2C on our environments, so we disabled it in our experiments. As in Algorithm~\ref{alg:moppo_nonshared},  the actor-to-critic ratio $C$ that we tune stands for the ratio $\eta_a/\eta_c$, where $\eta_a$ and $\eta_c$ are the learning rates of the actor and the critic optimizers, respectively. }
\label{alg:moa2c_nonshared}
\end{algorithm}

% \begin{algorithm}
% \DontPrintSemicolon
% \SetNoFillComment
% % \SetNlSty{}{}{}
% \SetKwInput{KwRequire}{Require}
% \KwRequire{Multi-objective MDP $\mathcal{M}=\left( \ssp,\A,\rw, P, \mu,\gamma \right)$}
% \KwRequire{Actor $\pi_\theta$, critic $\vc$ with parameters $\nu=\theta\cup\psi$}
% % Initialize $\nu$\;
% \While{not terminated}{
%     Sample reward weights $\balpha\sim U(\Delta_K)$\;
%     Sample a truncated trajectory from $P$ and $\pi_\theta(\cdot\mid \cdot,\balpha)$\;
%     \For(\tcp*[h]{iterate over epochs}){$e\gets 1$ \KwTo $E$ }{ 
%         Update PopArt statistics and the critic head \;
%         Get the reward-to-go and advantages using $\operatorname{TD}(1)$ \;
%         \For{$b\gets 1$ \KwTo $B$}{
%             Sample a minibatch\;
%             Update the entropy control gradient $\g$\;
%             Compute actor gradient $\nabla_\theta l_a$ and critic gradient $\nabla_\psi l_c$\;
%             Perform a joint update of the form $a\g+b\nabla_\theta l_a+c\nabla_\psi l_c$\;
%         }
%     }
% }
% % \Return{$\nu$}\;
% % \vspace{0.5em}
% \caption{The algorithm for the poster}
% \end{algorithm}

When the architectures of the actor and critic are joined, training becomes trickier. Now, multiple updates from the critic do not make much sense because this would unpredictably influence the actor. We want all updates of the actor parameters to be at least partially influenced by the relevant learning signal. Otherwise, we are in danger of a ``catastrophic misstep.''. Let now $\mathbf{g}_a$ be the actor gradient from the policy loss and the entropy regularization. Let $\mathbf{g}_c=\nabla_\psi l(\psi)$ be the gradient from the critic. Our intuition is that we need to make the norms of these gradients approximately proportional, i.e., $\norm{\mathbf{g}_c}\approx C \norm{\mathbf{g}_a}$, where $C>0$ is a hyperparameter. The most straightforward way to ensure this is to set $\beta_c=C\norm{\mathbf{g}_c}/\norm{\mathbf{g}_a}$. Setting $\beta_c$ to this value directly at each iteration would lead to a rather unstable learning rate for the critic, so instead, we compute a running average using a recurrent formula
\begin{equation}
  \beta_c = \frac{\delta C\norm{\mathbf{g}_c}}{\norm{\mathbf{g}_a}}+(1-\delta)\beta_c, \label{eq:dynbeta}
\end{equation}
where $\delta$ is another hyperparameter regulating the smoothness of the critic's learning rate, which we set to $0.001$ in all experiments. The hope is that the learning procedure is not as sensitive to $\delta$, while  $C$ is the analogue of the relative learning rate that we could compute in the separated actor and critic case.

We use Adam~\citep{kingma2014adam} for optimization after computing the above gradients. If the critic is separated, we naturally have a separate instance of Adam optimizer. Otherwise, the gradient estimates are fed into it after the adaptive relative weight update~\eqref{eq:dynbeta}.
\subsection{Step discarding heuristics}
During the preliminary experiments, we identified multiple scenarios that occur rather rarely but lead to the collapse of the learning process. Unstable training on toy environments was especially observed with the hypernetwork architecture. In part, it can be remedied by clipping the gradient norms (which we do) or even just by setting a lower learning rate, but this comes at the cost of lower sample efficiency. The collapse of learning that we observed mostly happened rapidly and led to trivial policies and zero entropy. To prevent the missteps that lead to collapse, we employ the following heuristics:
\begin{enumerate}
  \item If the entropy dropped significantly (the change in entropy is negative, and its absolute value is three standard deviations above the average absolute changes in entropy) \emph{and} the average reward did not increase from the previous step, we discard the step.
  \item If the entropy or actor gradient has been nearly zero for the past $200$ steps, reset learning from the last checkpoint where nonzero statistics were still observed.
\end{enumerate}
To prevent discarding too many steps, we set an upper limit of $5\%$ of the last  $100$ iterations that can be discarded.

\subsection{Entropy control}
\label{app:entropy}
The method described in Section~\ref{sec:entropy} to control entropy during training allows us to shape the entropy according to any schedule. An investigation into principled ways of selecting such schedules would be quite interesting, but it is out of the scope of this paper. We have selected three schedules based on our intuition about exploration-exploitation trade-offs. All of these schedules start from the maximal possible entropy $H_{target}(0)=H_{max}=\log |\A|$ and progress to $H_{target}(1)=H_{min}$. For Minecart and resource gathering experiments, we set $H_{min}=0.4$; for deep sea treasure, we set it to $0.1$. The schedules are defined as $H_{target}(u)$, where $u\in[0,1]$ denotes the proportion of environment interactions that were used so far from the total budget allocated for training.

The first schedule is a simple linear function going from the maximal entropy $H_{max}=\log |\A|$ to a desired minimal value $H_{min}$:
\begin{equation}
    H_{target,lin}(u)=H_{max}-(H_{max}-H_{min})u.
\end{equation}
The second schedule is normalized cosine, used to provide extra high-entropy exploration time in the beginning:
\begin{equation}
    H_{target,cos}(u)=(H_{max} - H_{min})\cos(\pi u/2)+H_{min}.
\end{equation}
Finally, we also designed a ``custom'' schedule with a flat start to provide exploration time and, unlike the previous schedule, has an extended flat end around the lower entropy to provide time for exploitation. We used the expression
\begin{equation}
    H_{target,custom}(u)=(H_{max} - H_{min})(0.5-\cos\left(\pi(1-u)^{1.3}\right)/2+H_{min}.
\end{equation}
The target entropy curves resulting from all of these methods are shown in Figure~\ref{fig:entropy} in the main paper. We also briefly experimented with ``resetting'' entropy to a high value a few times during training, similar to cosine learning rate annealing~\citep{loshchilov2016sgdr}, but did not get promising results.

\section{Hyperparameter tuning}
\label{app:hyper}
\subsection{Our methods}
\label{app:hyper_arch}
\begin{figure}
    \centering
    \captionsetup[subfigure]{justification=centering}
\begin{subfigure}{0.3\textwidth}
  \centering
  \includegraphics[width=\textwidth]{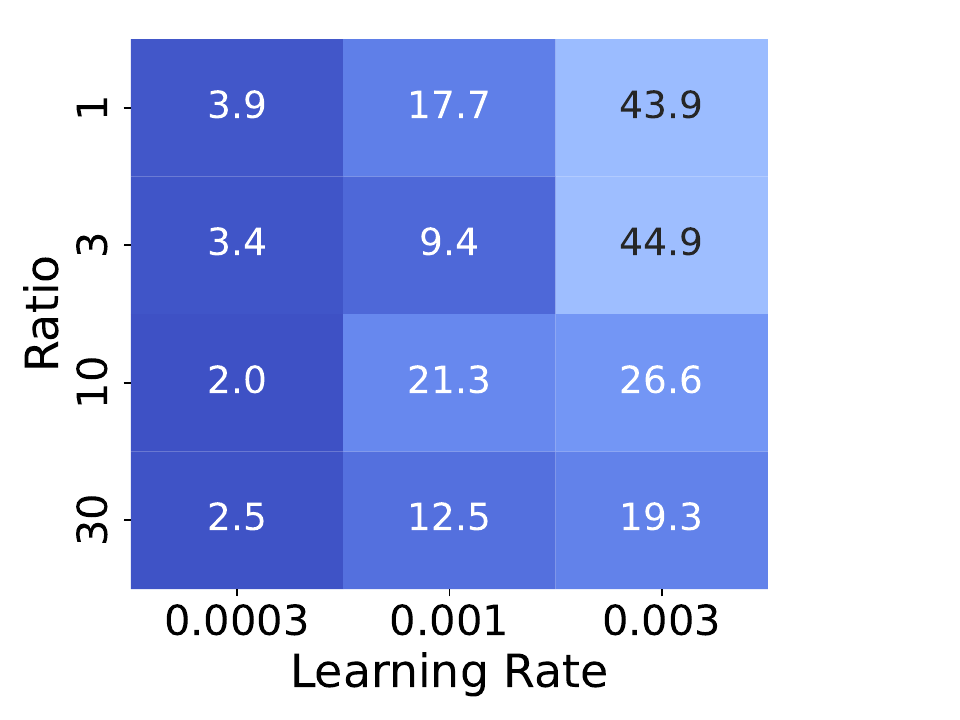 }
  \caption{ A2C, Shared Mult. embed., Fixed $\lambda=10^{-2}$ entropy }
\end{subfigure}
\hspace{0.1cm}
\begin{subfigure}{0.3\textwidth}
  \centering
  \includegraphics[width=\textwidth]{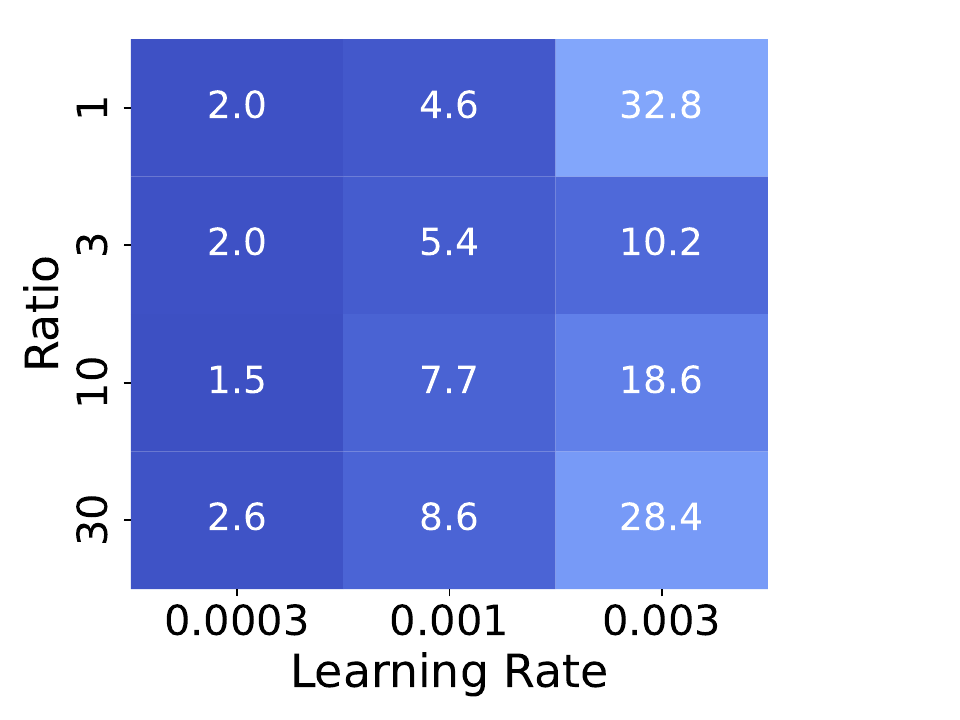 }
  \caption{ A2C, Non-shared Mult. embed., Fixed $\lambda=10^{-2}$ entropy }
\end{subfigure}
\hspace{0.1cm}
\begin{subfigure}{0.3\textwidth}
  \centering
  \includegraphics[width=\textwidth]{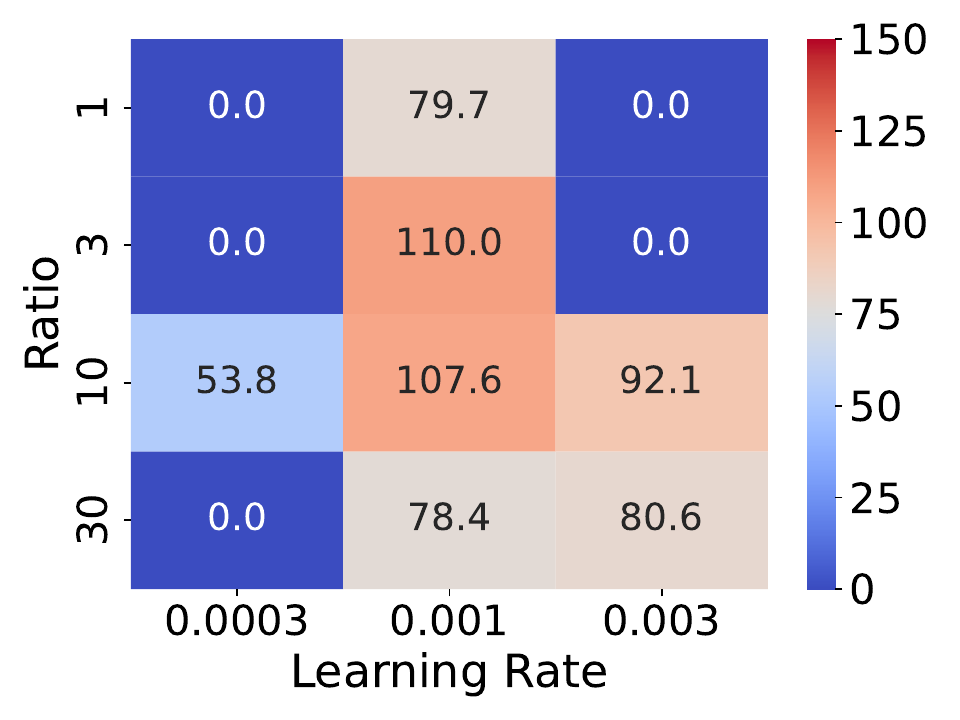 }
  \caption{ PPO, Shared Mult. embed., Custom entropy }
\end{subfigure}

\vspace{1em}

\begin{subfigure}{0.3\textwidth}
  \centering
  \includegraphics[width=\textwidth]{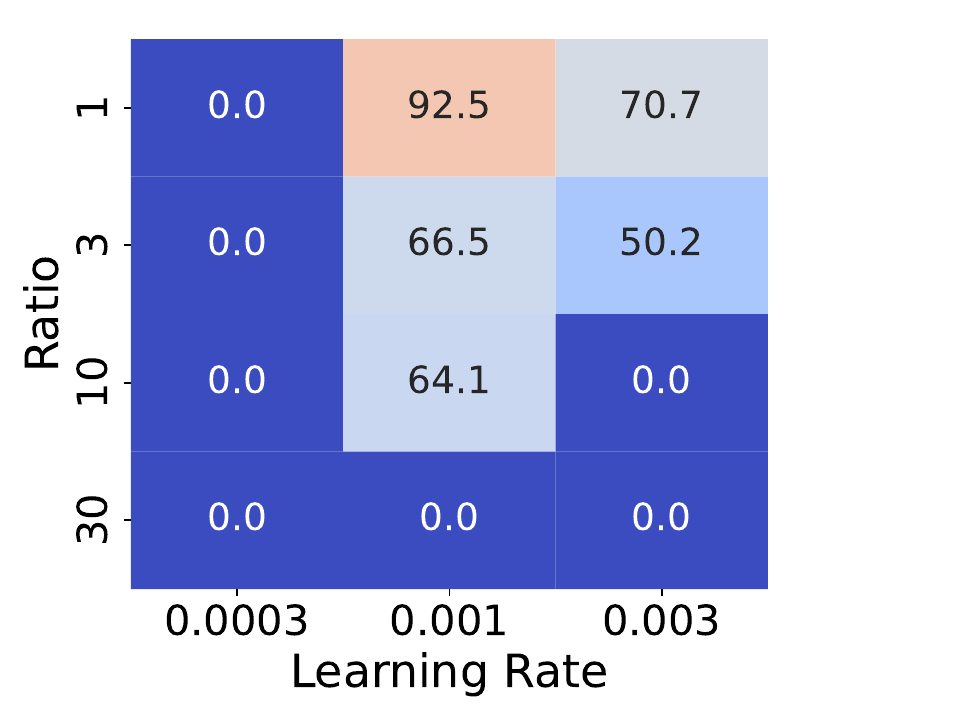 }
  \caption{ PPO, Non-shared Mult. embed., Custom entropy }
\end{subfigure}
\hspace{0.1cm}
\begin{subfigure}{0.3\textwidth}
  \centering
  \includegraphics[width=\textwidth]{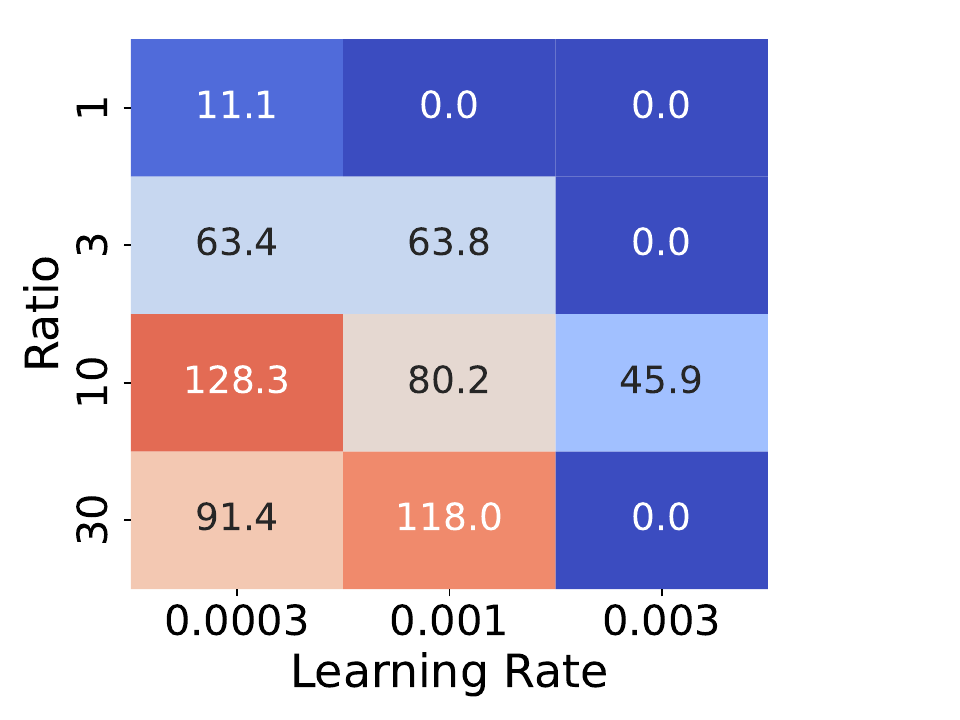 }
  \caption{ PPO, Shared Hypernet, Custom entropy, No PopArt }
\end{subfigure}\hspace{0.1cm}
\begin{subfigure}{0.3\textwidth}
  \centering
  \includegraphics[width=\textwidth]{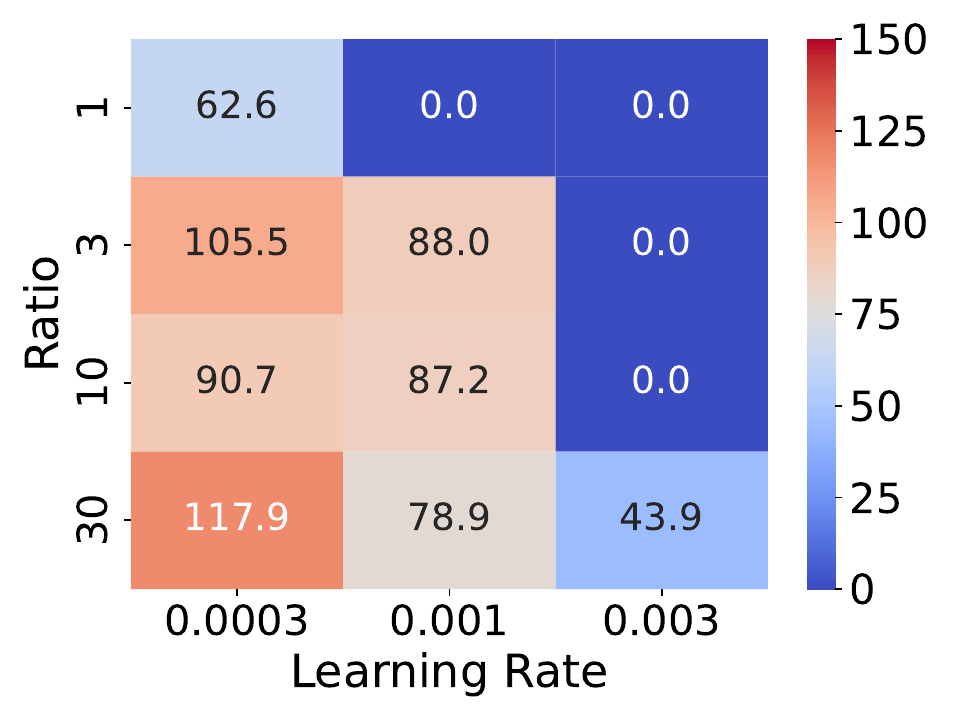 }
  \caption{ PPO, Shared Hypernet+o., Custom entropy, No PopArt }
\end{subfigure}

\vspace{1em}

\begin{subfigure}{0.3\textwidth}
  \centering
  \includegraphics[width=\textwidth]{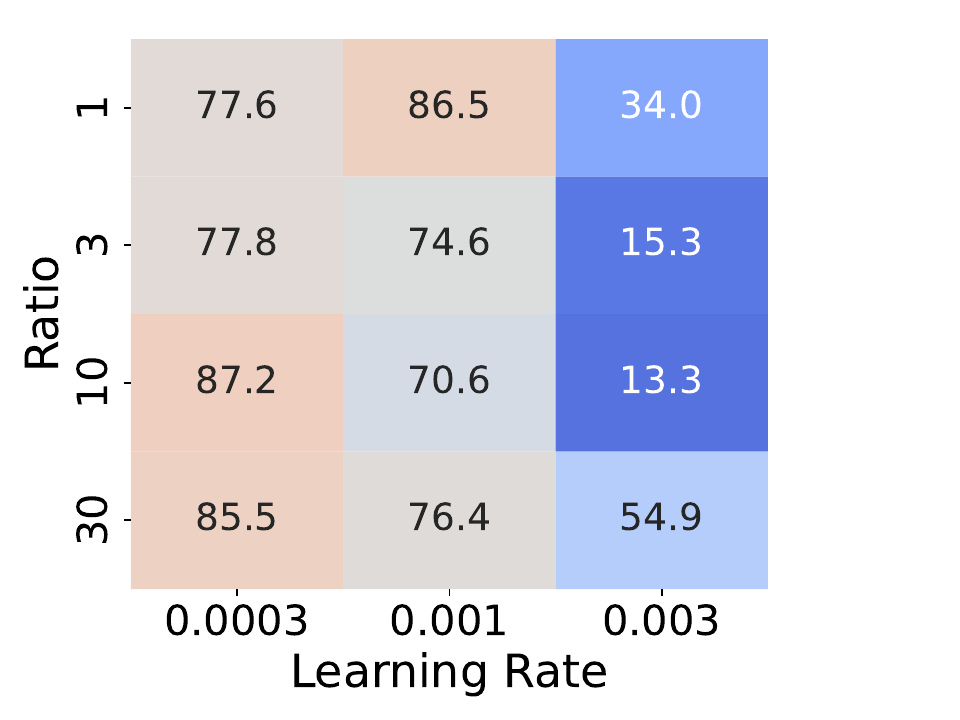 }
  \caption{ PPO, Non-shared Hypernet, Custom entropy, No PopArt }
\end{subfigure}
\hspace{0.1cm}
\begin{subfigure}{0.3\textwidth}
  \centering
  \includegraphics[width=\textwidth]{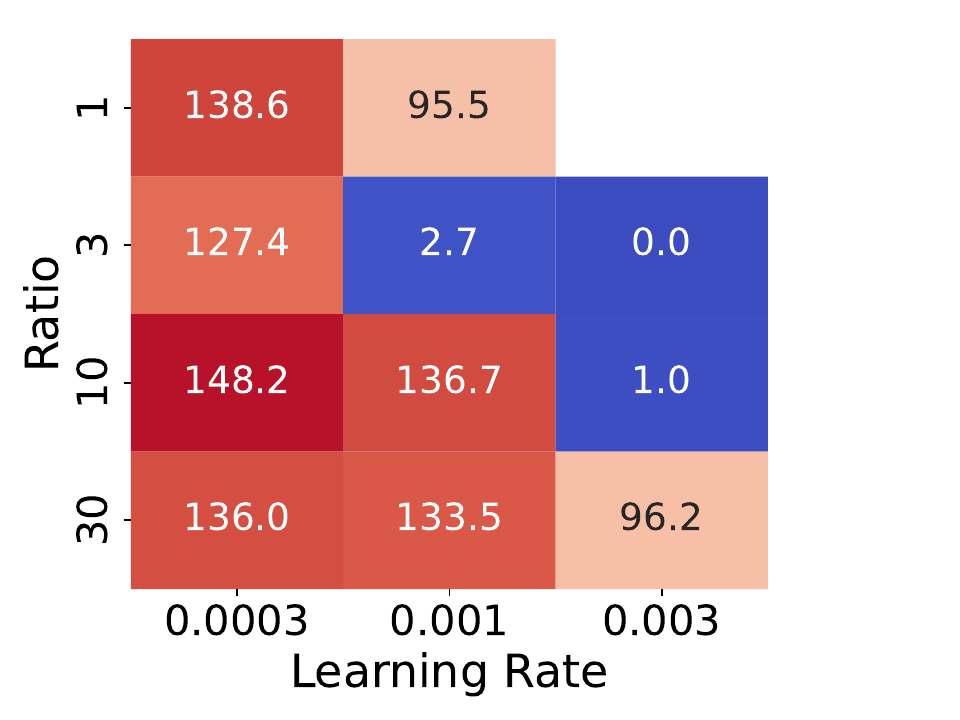 }
  \caption{ PPO, Shared Multi-body, Custom entropy }
\end{subfigure}
\hspace{0.1cm}
\begin{subfigure}{0.3\textwidth}
  \centering
  \includegraphics[width=\textwidth]{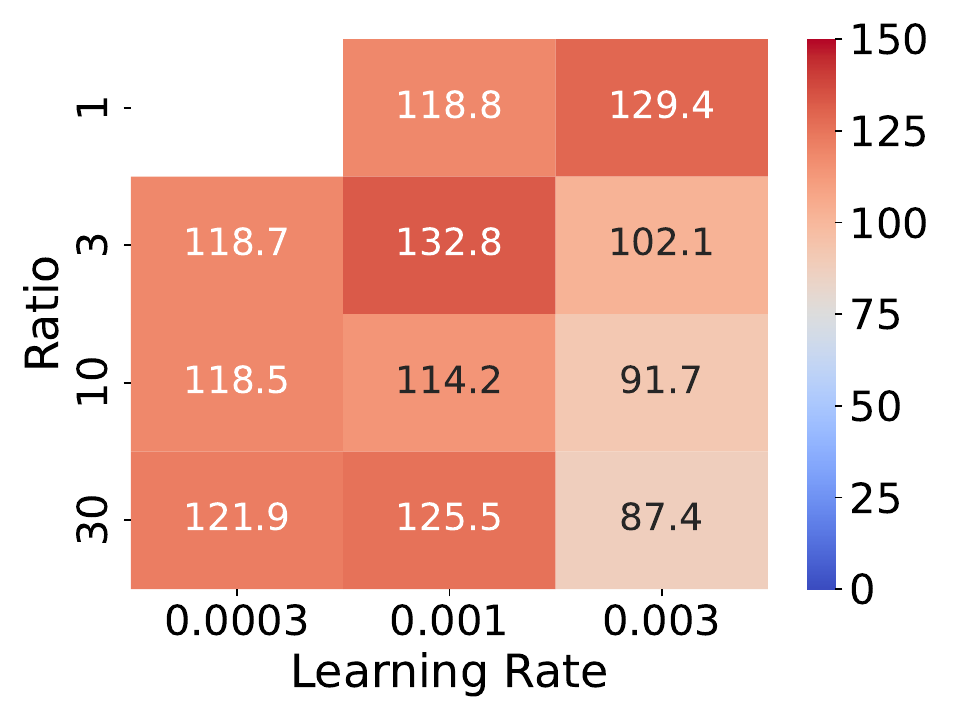 }
  \caption{ PPO, Non-shared Multi-body, Custom entropy }
\end{subfigure}
    \caption{Hypervolume of all hyperparameter configurations for the ablations demonstrated in Figure~\ref{fig:minecart-tuning} and Table~\ref{tab:big_table} run on the non-deterministic Minecart environment. Part 1. ``Hypernet+o.'' is the architecture where the observation is provided to the hypernetwork along with $\boldsymbol{\alpha}$.}
    \label{fig:hyperparam_stoch_1}
\end{figure}

\begin{figure}[htbp!]
    \centering
    \captionsetup[subfigure]{justification=centering}
\begin{subfigure}{0.3\textwidth}
  \centering
  \includegraphics[width=\textwidth]{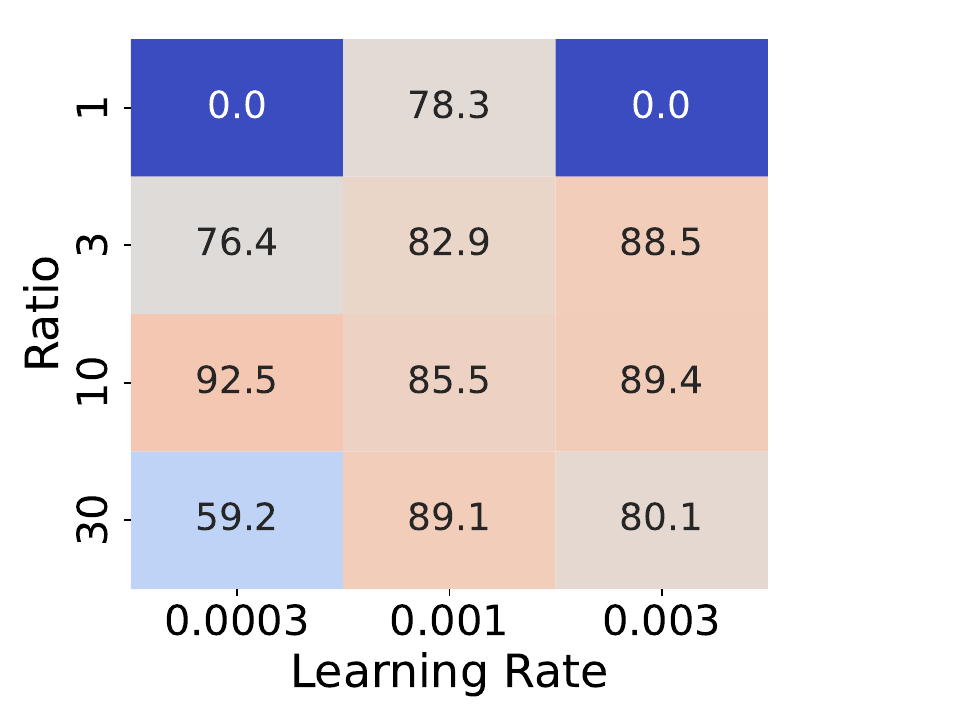 }
  \caption{ PPO, Shared Mult. embed., Cosine entropy }
\end{subfigure}
\hspace{0.1cm}
\begin{subfigure}{0.3\textwidth}
  \centering
  \includegraphics[width=\textwidth]{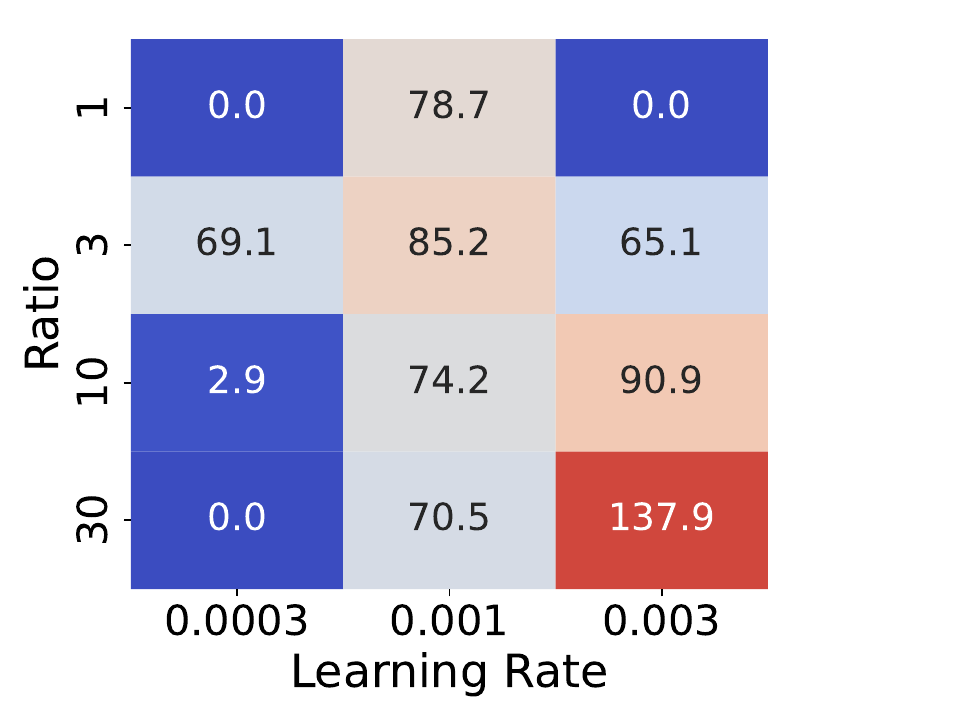 }
  \caption{ PPO, Shared Mult. embed., Linear entropy }
\end{subfigure}
\hspace{0.1cm}
\begin{subfigure}{0.3\textwidth}
  \centering
  \includegraphics[width=\textwidth]{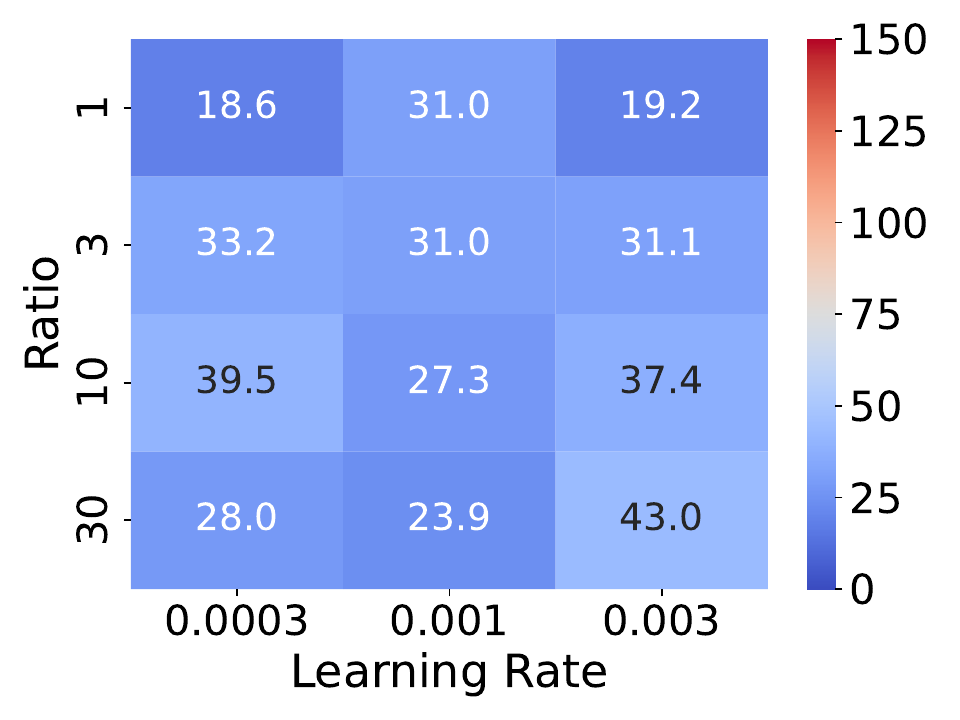 }
  \caption{ PPO, Shared Mult. embed., Fixed $\lambda=10^{-1}$ entropy }
\end{subfigure}

\vspace{1em}

\begin{subfigure}{0.3\textwidth}
  \centering
  \includegraphics[width=\textwidth]{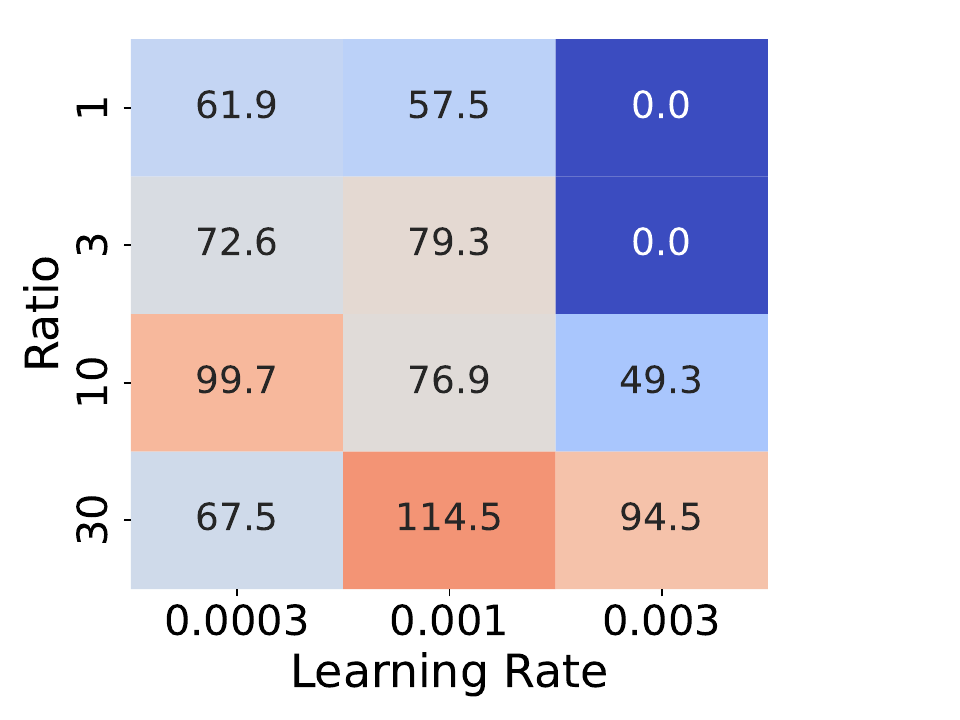 }
  \caption{ PPO, Shared Mult. embed., Fixed $\lambda=10^{-2}$ entropy }
\end{subfigure}
\hspace{0.1cm}
\begin{subfigure}{0.3\textwidth}
  \centering
  \includegraphics[width=\textwidth]{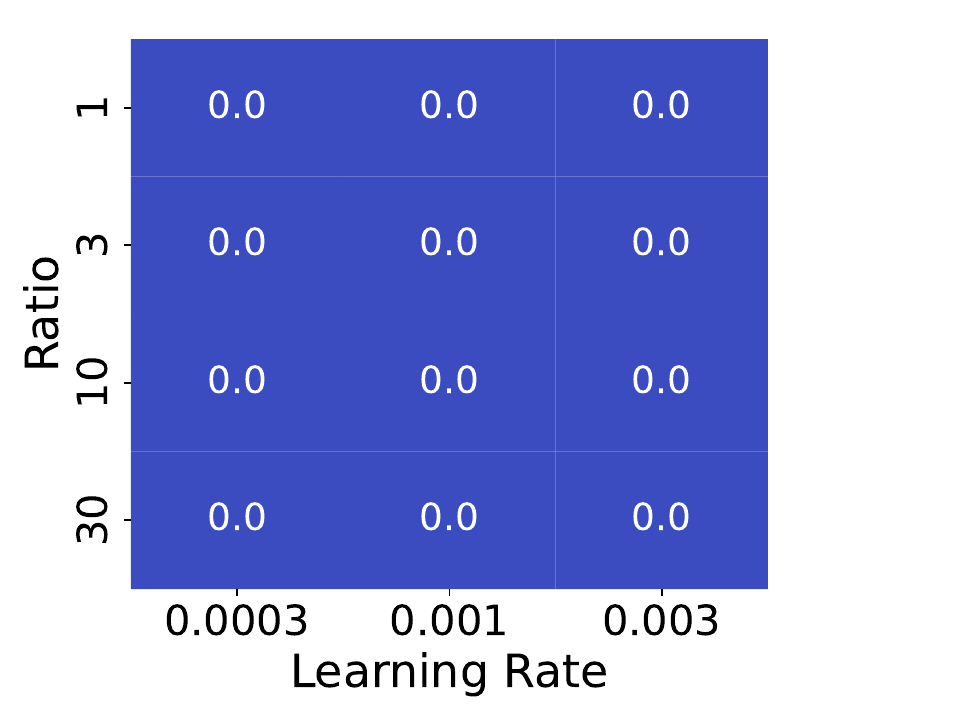 }
  \caption{ PPO, Shared Mult. embed., Fixed $\lambda=10^{-3}$ entropy }
\end{subfigure}
\hspace{0.1cm}
\begin{subfigure}{0.3\textwidth}
  \centering
  \includegraphics[width=\textwidth]{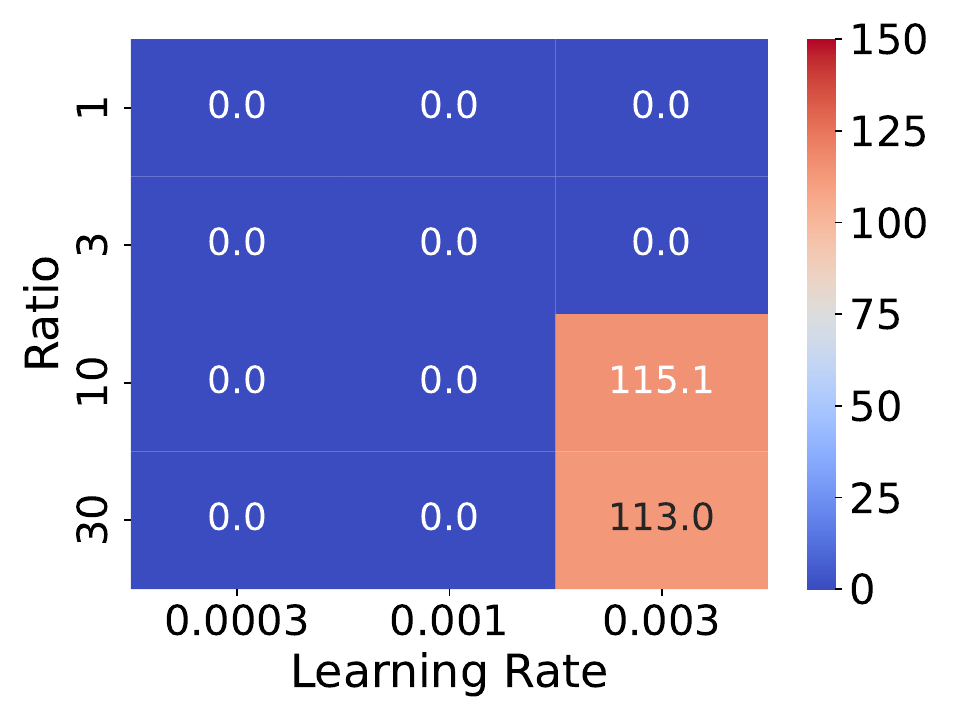 }
  \caption{ PPO, Shared Mult. embed., Custom entropy, No renorm. }
\end{subfigure}

\vspace{1em}

\begin{subfigure}{0.3\textwidth}
  \centering
  \includegraphics[width=\textwidth]{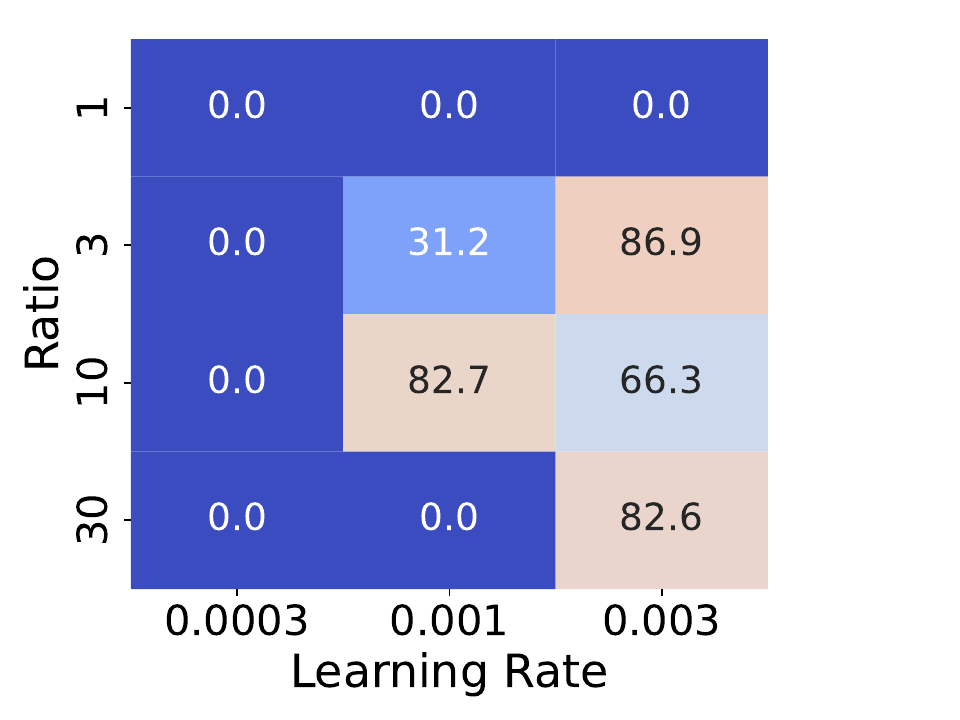 }
  \caption{ PPO, Shared Mult. embed., Custom entropy, No PopArt }
\end{subfigure}
\hspace{0.1cm}
\begin{subfigure}{0.3\textwidth}
  \centering
  \includegraphics[width=\textwidth]{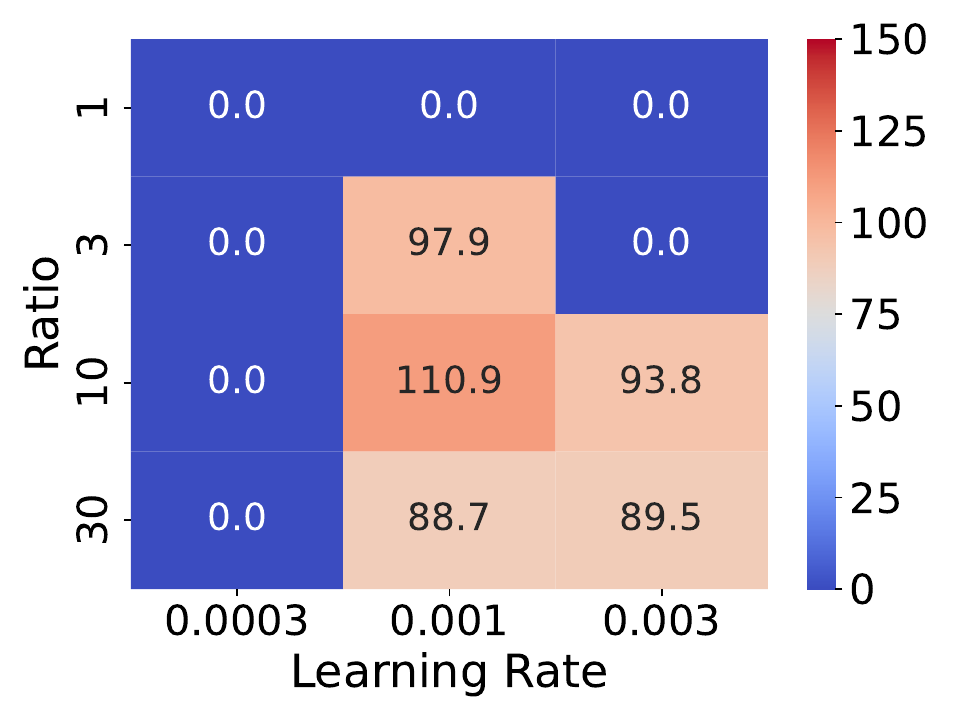 }
  \caption{ PPO, Shared Mult. embed., Custom entropy, No dyn. beta }
\end{subfigure}
    \caption{Hypervolume of all hyperparameter configurations for the ablations demonstrated in Figure~\ref{fig:minecart-tuning} and Table~\ref{tab:big_table} run on the non-deterministic Minecart environment. Part 2.}
    \label{fig:hyperparam_stoch_2}
\end{figure}

Here we describe the details of the hyperparameter search that we performed for each architecture when comparing them to produce Figure~\ref{fig:minecart-tuning} in Section~\ref{sec:hyperparams}. We run a hyperparameter search for all shared trunk architectures and pick the best final hypervolume. We search over the learning rates $\eta\in$\{3e-4, 1e-3, 3e-3\} and over the ratio $C\in\{1,3,10,30\}$ as defined in Algorithm~\ref{alg:moppo}. For architectures with non-shared trunks, we instead search over the learning rate of the critic among  $\eta_c\in \{\eta,\eta/3,\eta/10,\eta/30\}$. All hyperparameter grids for this experiment are shown in Figures~\ref{fig:hyperparam_stoch_1} and~\ref{fig:hyperparam_stoch_2}. Final results on more architectures with extra metrics for the Pareto front are shown in Table~\ref{tab:big_table}.

For experiments in Section~\ref{sec:comp_baselines}, we also run a hyperparameter grid search for each pair of methods and environments. For the reacher environment, we search over $\eta\in$\{3e-4, 1e-3, 3e-3\} and $C\in\{0.33,1,3\}$. We picked smaller values of $C$ for this environment because the trajectories sampled from it are always infinite; hence, bootstrapping and the critic are more relevant to the algorithm's performance.

\subsection{PCN and Envelope Q-learning}
\label{app:hyper_baselines}

We aim to compare our methods fairly to PCN by performing additional hyperparameter sweeps for PCN. We sweep through the same learning rates $\eta\in\left\{0.0001,0.0003,0.001,0.003 \right\}$ and perform an architecture search. The PCN network consists of embedding layers for the target rewards and for the state, followed by an MLP that combines these into the final action. We experiment with three sizes of the hidden layer of the MLP: $d\in\{64, 128, 256\}$. In addition, we try to add an extra hidden layer. This makes the total number of checked configurations $4\times 3\times 2=24$.

The default implementation of Envelope Q-learning already relies on a rather deep MLP ($4$ hidden layers with width $256$ each). We expected less sensitivity to the architecture of the MLP than to ways in which one provides the state and $\balpha$ to it (our reference implmentation simply concatenates them), so we instead focused on the $\epsilon$ for exploration and the learning rate. The reference implementation linearly anneals the exploration $\epsilon$ from $1$ to some value $\epsilon_{min}$ in the first half of the training, and then keeps it constant. For Envelope Q-learning, we sweeped through the learning rates $\eta\in\{0.0003, 0.001, 0.003\}$ and through $\epsilon_{min}\in\{0.01, 0.05, 0.1\}$. Other parameters of the baselines were kept at the default values from MORL-Baselines.

\section{Additional Results}
\label{app:extra_results}
In Table~\ref{tab:extra_hyperparam}, we show the hyperparameters of our algorithms that were not tuned via a grid search.

\begin{table}[t]
    \centering
    \begin{tabular}{c||c|c|c}
         & Deep-sea-treasure & Minecart & Reacher \\
        \hline
        \hline
       Hidden dim. for merge networks  & \multicolumn{3}{c}{256} \\
       \hline
       Hidden dim. for multi-body networks  & \multicolumn{3}{c}{256} \\
       \hline
       Hidden dim. for hypernetworks  & \multicolumn{3}{c}{64} \\
       \hline
       $\gamma$ & \multicolumn{3}{c}{0.99} \\
       \hline
       Initial $\lambda$ for dynamic entropy & \multicolumn{3}{c}{0.01} \\
       \hline
       $H_{min}$ for entropy control & \multicolumn{3}{c}{0.4} \\
       \hline
       Dampening $c$ for entropy control & \multicolumn{3}{c}{0.01} \\
       \hline
       Total number of environment interactions & 1e5 & 4e6 & 1e6 \\
       \hline
       Number of trajectories $B$ in a batch & \multicolumn{3}{c}{8} \\
       \hline
       Number of PPO epochs & \multicolumn{3}{c}{4} \\
       \hline
       Number of PPO minibatches & \multicolumn{3}{c}{8} \\
       \hline
       PPO clip $\varepsilon$ & \multicolumn{3}{c}{0.2} \\
       \hline
       PopArt learning rate & \multicolumn{3}{c}{0.001} \\
       \hline
       Critic weight decay & \multicolumn{3}{c}{0.01} \\
       \hline
       Max gradient norm for the actor & \multicolumn{3}{c}{0.5} \\
       \hline
       Max gradient norm for the critic & \multicolumn{3}{c}{0.5} \\
       \hline
       Hypervolume reference & --- & (0,0,-200) & (-50,-50,-50,-50)
    \end{tabular}
    \caption{Hyperparameter choices for our experiments. Notice that we did not tune any hyperparameters other than the number of samples for individual environments, apart from the learning rate $\eta$ and the learning rate ratio $C$ as discussed in the main paper.}
    \label{tab:extra_hyperparam}
\end{table}

\subsection{Ablations on normalization heuristics}
\label{app:normalization_ablation}
Table~\ref{tab:big_table}, apart from the methods discussed in Section~\ref{sec:hyperparams}, shows three ablations on parts of our approach. The results include:
\begin{itemize}
    \item The dynamic gradient reweighting approach for shared trunk architectures described in Appendix~\ref{app:combining} does not provide benefits compared to using a constant relative weight. Although we believe this approach might stabilize learning dynamics in some other environments, we chose not to describe it in the main paper because of this result.
    \item If we do not use the normailzed scalarization~\eqref{eq:PopArt_scalarization}, the performance slightly improves. Note, however, that this scalarization makes the approach scale-invariant, so the minor performance loss is probably justified.
    
\end{itemize}

\begin{table}[t]
    \centering
    \resizebox{\textwidth}{!}{
    \begin{tabular}{cccccccccc}
\multicolumn{5}{c}{Method} && \multicolumn{3}{c}{Metrics on Minecart} \\
\cmidrule{1-5} \cmidrule{7-9}
Algo & Arch & Shared trunk & Entropy & Notes && HV & EU & MUL \\
\hline
\hline
A2C & Merge net. & Yes & Fixed $\lambda=10^{-2}$ & --- && $28.3 \pm 25.7$ & $-0.34 \pm 0.21$ & $0.99 \pm 0.21$  \\
A2C & Merge net. & No & Fixed $\lambda=10^{-2}$ & --- && $41.5 \pm 37.5$ & $-0.24 \pm 0.25$ & $0.81 \pm 0.25$  \\
PPO & Merge net. & Yes & Custom & --- && $76.7 \pm 22.3$ & $-0.03 \pm 0.18$ & $0.59 \pm 0.18$  \\
PPO & Merge net. & No & Custom & --- && $38.0 \pm 43.0$ & $-0.11 \pm 0.10$ & $0.82 \pm 0.10$  \\
PPO & Hypernet w/ obs. & Yes & Custom & No PopArt && $107.7 \pm 18.9$ & $0.22 \pm 0.05$ & $0.25 \pm 0.05$  \\
PPO & Hypernet & Yes & Custom & No PopArt && $100.1 \pm 15.5$ & $0.19 \pm 0.03$ & $0.31 \pm 0.03$  \\
PPO & Hypernet & No & Custom & No PopArt && $87.1 \pm 2.9$ & $0.14 \pm 0.01$ & $0.33 \pm 0.01$  \\
PPO & Multi-body & Yes & Custom & --- && $134.7 \pm 11.3$ & $0.27 \pm 0.01$ & $0.16 \pm 0.01$  \\
PPO & Multi-body & No & Custom & --- && $118.1 \pm 16.7$ & $0.18 \pm 0.05$ & $0.34 \pm 0.05$  \\
PPO & Merge net. & Yes & Cos & --- && $74.1 \pm 14.2$ & $-0.03 \pm 0.07$ & $0.56 \pm 0.07$  \\
PPO & Merge net. & Yes & Linear & --- && $81.1 \pm 44.2$ & $-0.06 \pm 0.31$ & $0.64 \pm 0.31$  \\
PPO & Merge net. & Yes & Fixed $\lambda=10^{-1}$ & --- && $41.2 \pm 4.5$ & $-0.42 \pm 0.08$ & $1.05 \pm 0.08$  \\
PPO & Merge net. & Yes & Fixed $\lambda=10^{-2}$ & --- && $84.4 \pm 33.1$ & $-0.00 \pm 0.10$ & $0.61 \pm 0.10$  \\
PPO & Merge net. & Yes & Fixed $\lambda=10^{-3}$ & --- && $0.3 \pm 0.6$ & $-0.65 \pm 0.09$ & $1.30 \pm 0.09$  \\
PPO & Merge net. & Yes & Custom & No dyn. beta && $82.2 \pm 20.9$ & $0.04 \pm 0.05$ & $0.56 \pm 0.05$  \\
PPO & Merge net. & Yes & Custom & No renorm. && $92.4 \pm 21.4$ & $0.16 \pm 0.05$ & $0.37 \pm 0.05$  \\
PPO & Merge net. & Yes & Custom & No PopArt && $59.5 \pm 33.9$ & $0.08 \pm 0.09$ & $0.50 \pm 0.09$  \\
    \end{tabular}
    }
    \caption{Detailed results of ablation studies on a stochastic Minecart and resource gathering. ``No renorm.'' means that PopArt was used for value learning, but normalized scalarization~\eqref{eq:PopArt_scalarization} was not used when computing the policy loss. ``No dyn. beta'' means that the relative gradient weighting scheme from Appendix~\ref{app:combining}. HV is hypervolume (larger is better), EU is expected utility (larger is better), and MUL is maximum utility loss (smaller is better). Negative MUL can occur due to a stochastic environment, which allows the sampled reward to be higher than the expected optimal reward. }
    \label{tab:big_table}
\end{table}

\section{Compute requirements}
\label{app:compute}
We used a cluster with Nvidia V100 GPU nodes. An average run on MO-reacher took around $3$ hours, and we ran $3$ of them in parallel. Hence, reproducing the results on MO-reacher would take around 75 GPU hours. A run on minecart took around $5$ hours, and we also ran $3$ in parallel. In total this amounds to ~490 GPU hours for minecart. Hence, to reproduce the results from the main paper, it takes roughly $565$ hours on a V100 GPU.

% LaTeX for subfigures was generated

\begin{figure}[htbp!]
    \centering
    \captionsetup[subfigure]{justification=centering}
\begin{subfigure}{0.3\textwidth}
  \centering
  \includegraphics[width=\textwidth]{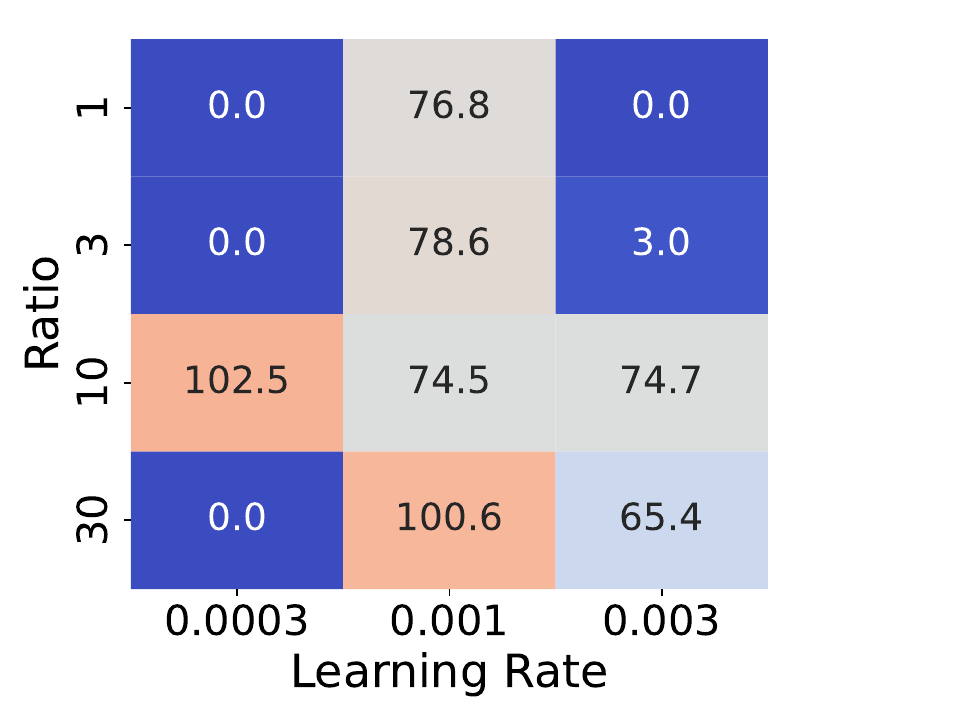 }
  \caption{ PPO, Shared Mult. embed., Custom entropy }
\end{subfigure}
\hspace{0.1cm}
\begin{subfigure}{0.3\textwidth}
  \centering
  \includegraphics[width=\textwidth]{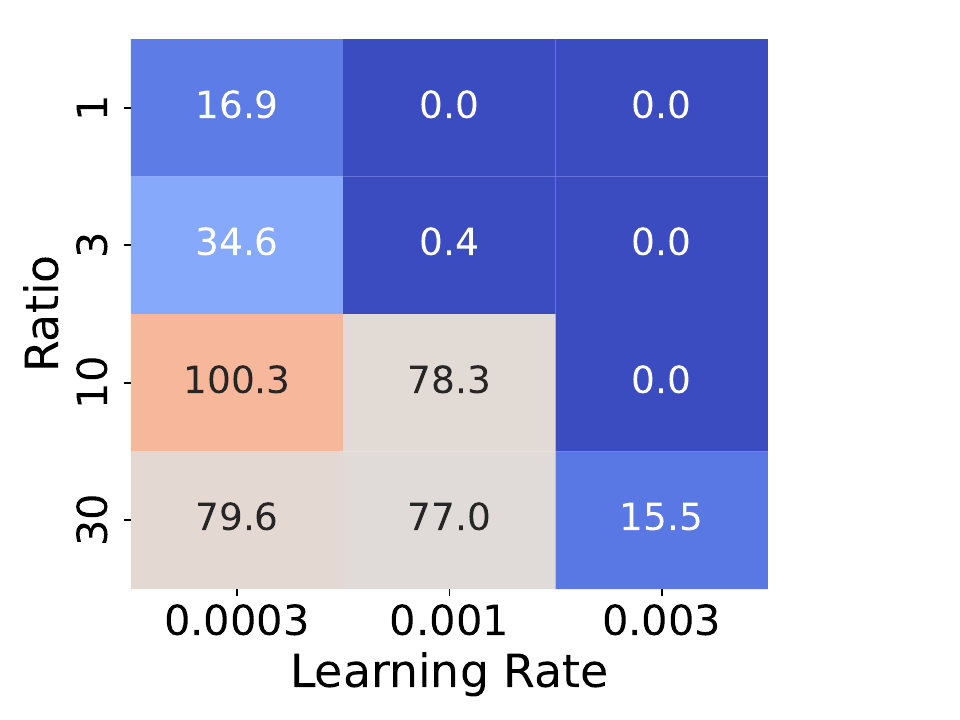 }
  \caption{ PPO, Shared Hypernet, Custom entropy, No PopArt }
\end{subfigure}
\hspace{0.1cm}
\begin{subfigure}{0.3\textwidth}
  \centering
  \includegraphics[width=\textwidth]{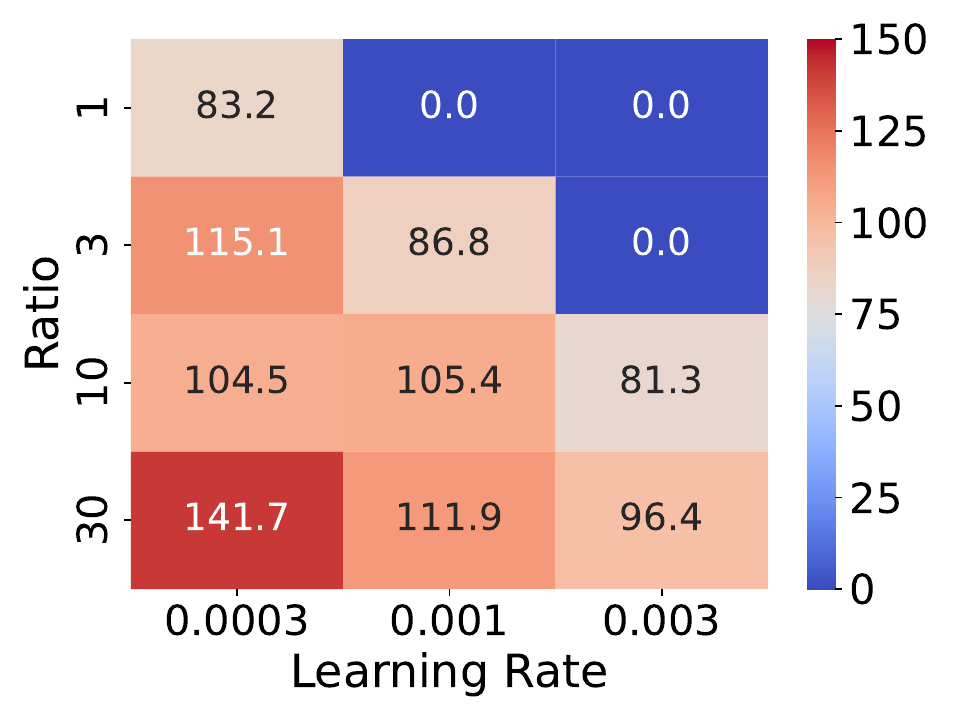 }
  \caption{ PPO, Shared Multi-body, Custom entropy }
\end{subfigure}

\vspace{1em}

\begin{subfigure}{0.3\textwidth}
  \centering
  \includegraphics[width=\textwidth]{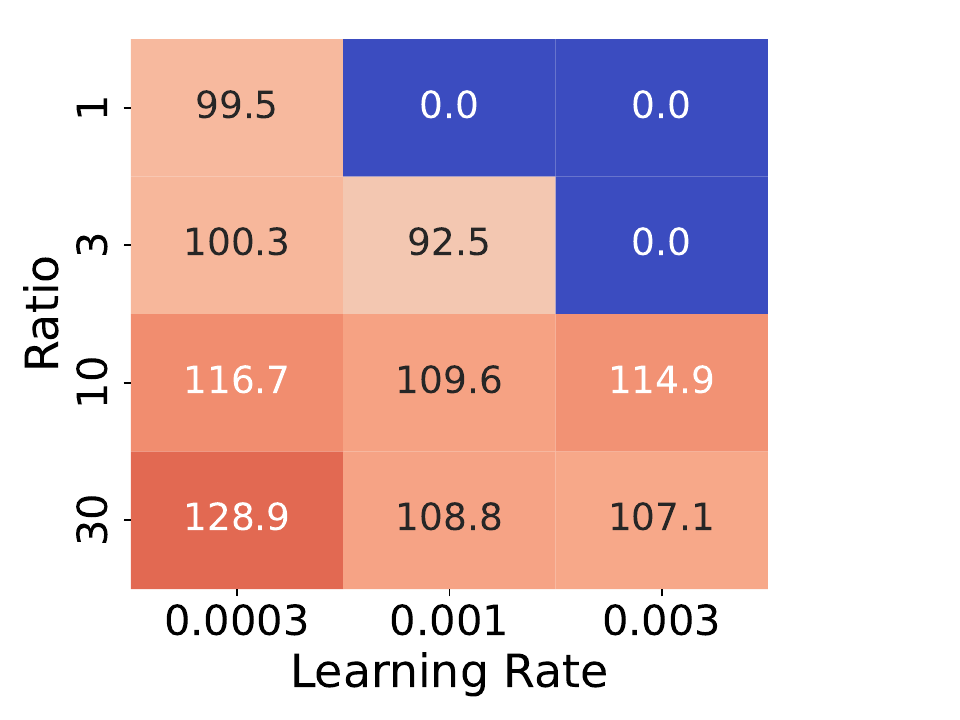 }
  \caption{ PPO, Shared Multi-body, Linear entropy }
\end{subfigure}
\hspace{0.1cm}
\begin{subfigure}{0.3\textwidth}
  \centering
  \includegraphics[width=\textwidth]{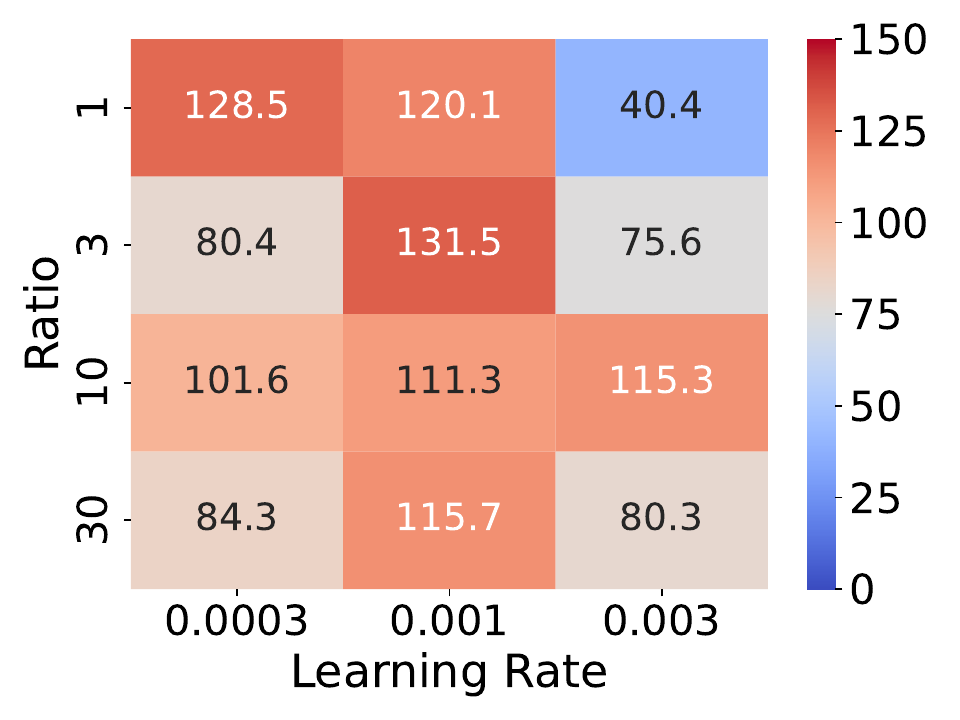 }
  \caption{ PPO, Non-shared Multi-body, Custom entropy }
\end{subfigure}

    \caption{Hypervolume of our hyperparameter configurations for the ablations demonstrated in Table~\ref{tab:baselines} run on the deterministic Minecart environment.}
    \label{fig:hyperparam_det}
\end{figure}

\begin{figure}[htbp!]
    \centering
    \captionsetup[subfigure]{justification=centering}
\begin{subfigure}{0.3\textwidth}
  \centering
  \includegraphics[width=\textwidth]{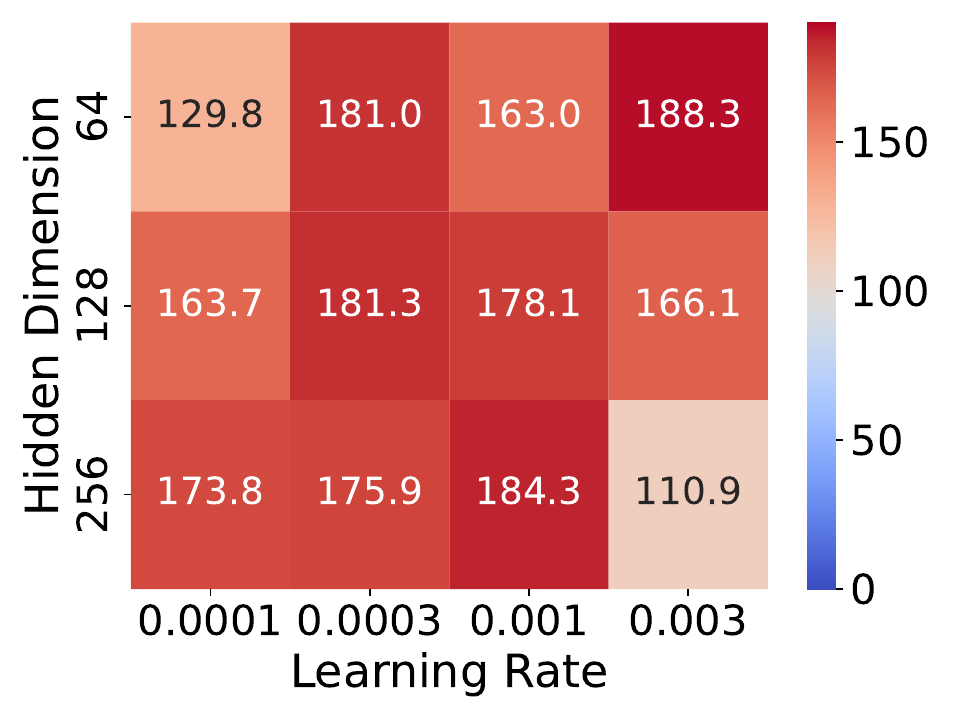 }
  \caption{ PCN, deep net, minecart det }
\end{subfigure}
\hspace{0.1cm}
\begin{subfigure}{0.3\textwidth}
  \centering
  \includegraphics[width=\textwidth]{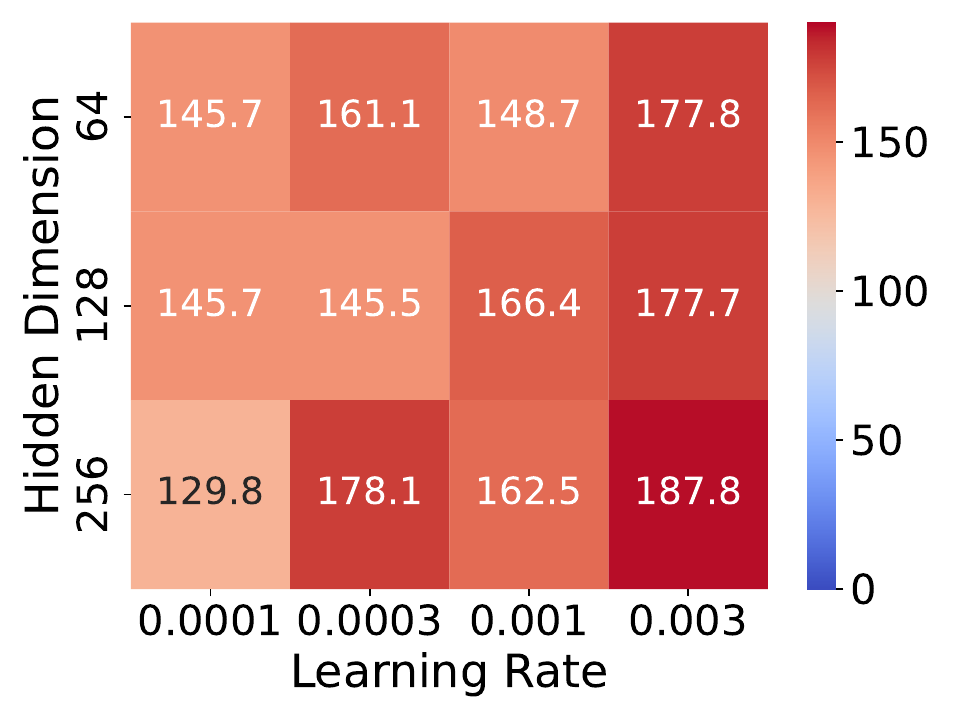 }
  \caption{ PCN, shallow net, minecart det }
\end{subfigure}

\begin{subfigure}{0.3\textwidth}
  \centering
  \includegraphics[width=\textwidth]{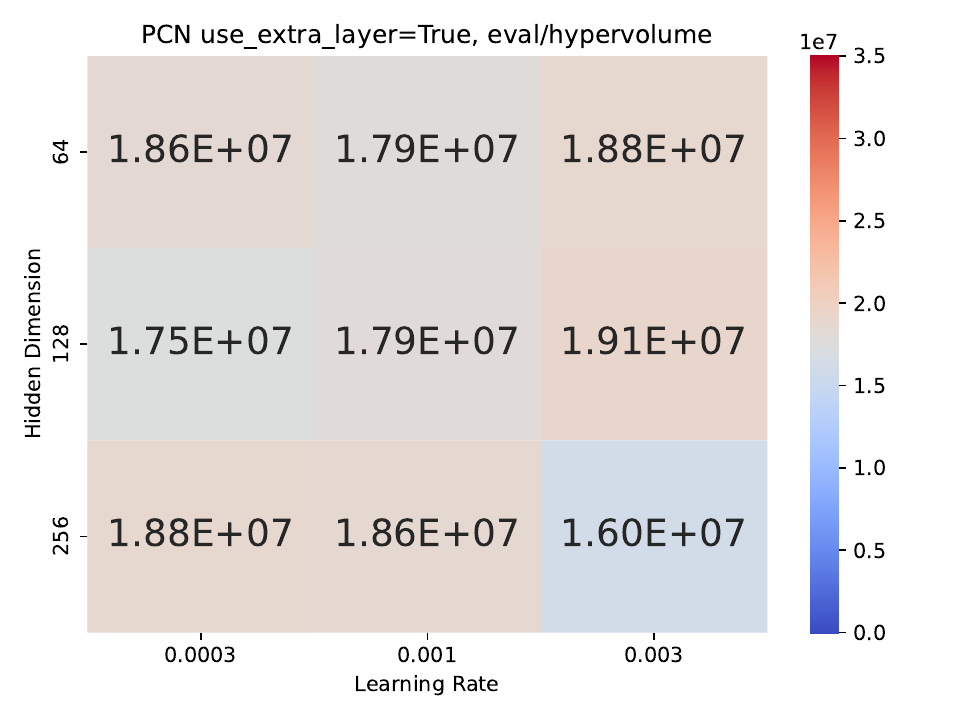 }
  \caption{ PCN, deep net, Reacher }
\end{subfigure}
\hspace{0.1cm}
\begin{subfigure}{0.3\textwidth}
  \centering
  \includegraphics[width=\textwidth]{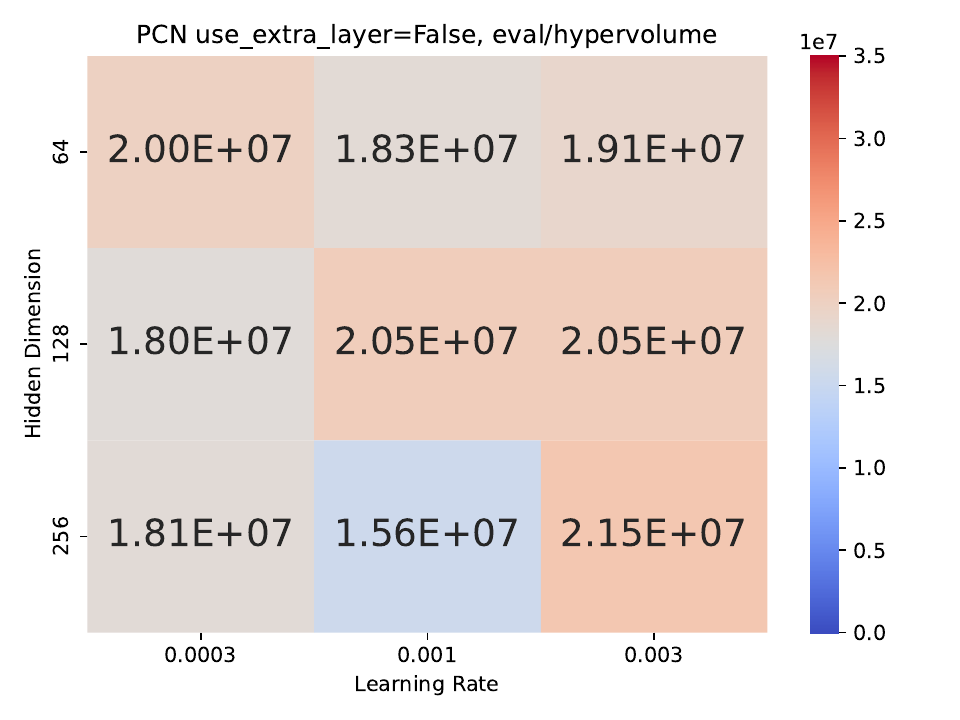 }
  \caption{ PCN, shallow net, Reacher }
\end{subfigure}
    \caption{Hypervolume of all hyperparameter configurations for the PCN networks on minecart-deterministic and on Reacher.}
    \label{fig:hyperparam_pcn}
\end{figure}

\begin{figure}[htbp!]
    \centering
    \captionsetup[subfigure]{justification=centering}
\begin{subfigure}{0.3\textwidth}
  \centering
  \includegraphics[width=\textwidth]{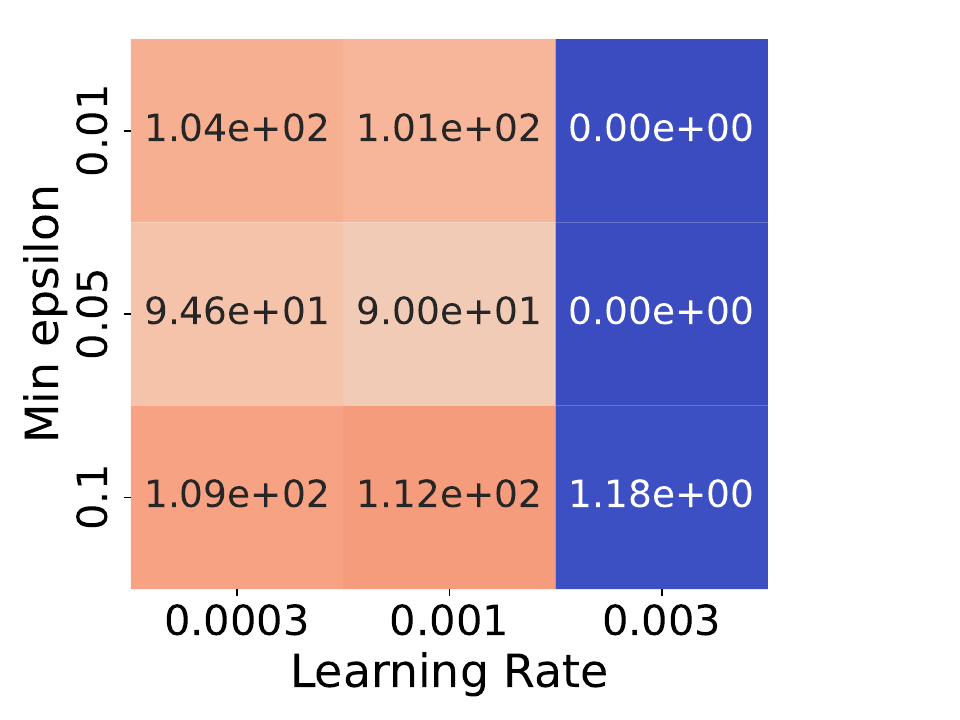}
  \caption{ Minecart }
\end{subfigure}
\hspace{0.1cm}
\begin{subfigure}{0.3\textwidth}
  \centering
  \includegraphics[width=\textwidth]{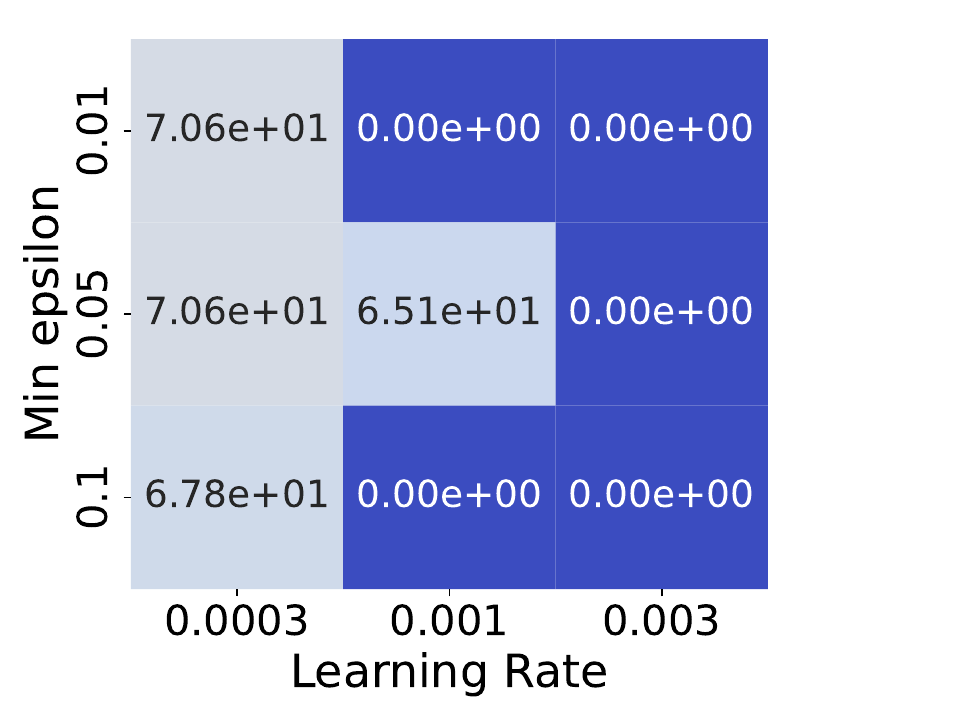}
  \caption{ Minecart det. }
\end{subfigure}
\hspace{0.1cm}
\begin{subfigure}{0.3\textwidth}
  \centering
  \includegraphics[width=\textwidth]{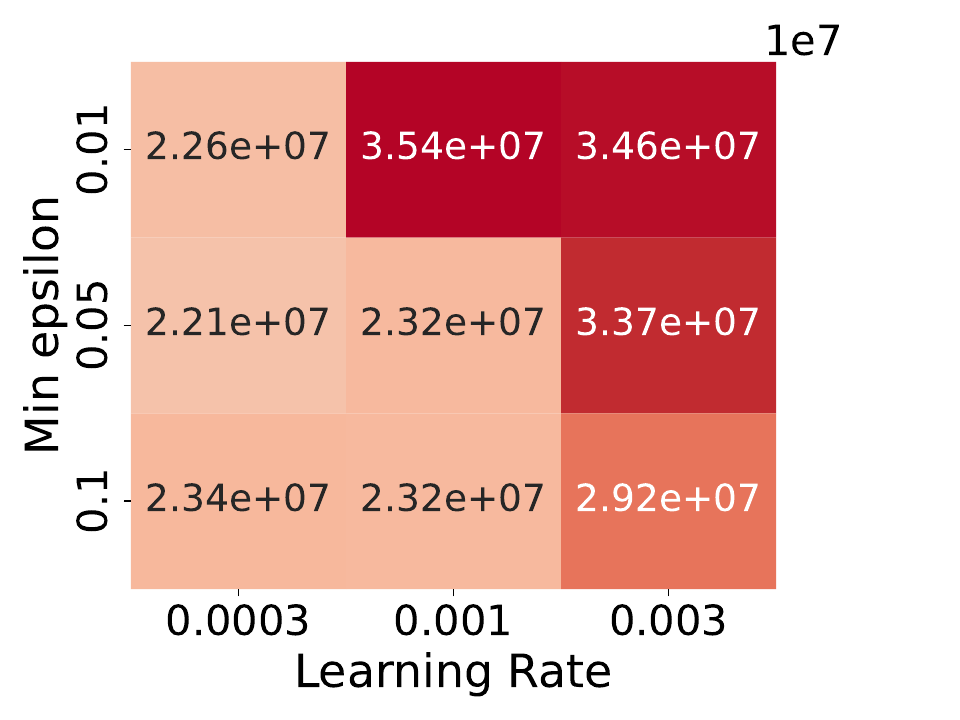}
  \caption{ Reacher }
\end{subfigure}
    \caption{Hypervolume of all hyperparameter configurations for Envelope Q-learning evaluation.}
    \label{fig:hyperparam_envelope}
\end{figure}

\begin{figure}[htbp!]
    \centering
    \captionsetup[subfigure]{justification=centering}
\begin{subfigure}{0.3\textwidth}
  \centering
  \includegraphics[width=\textwidth]{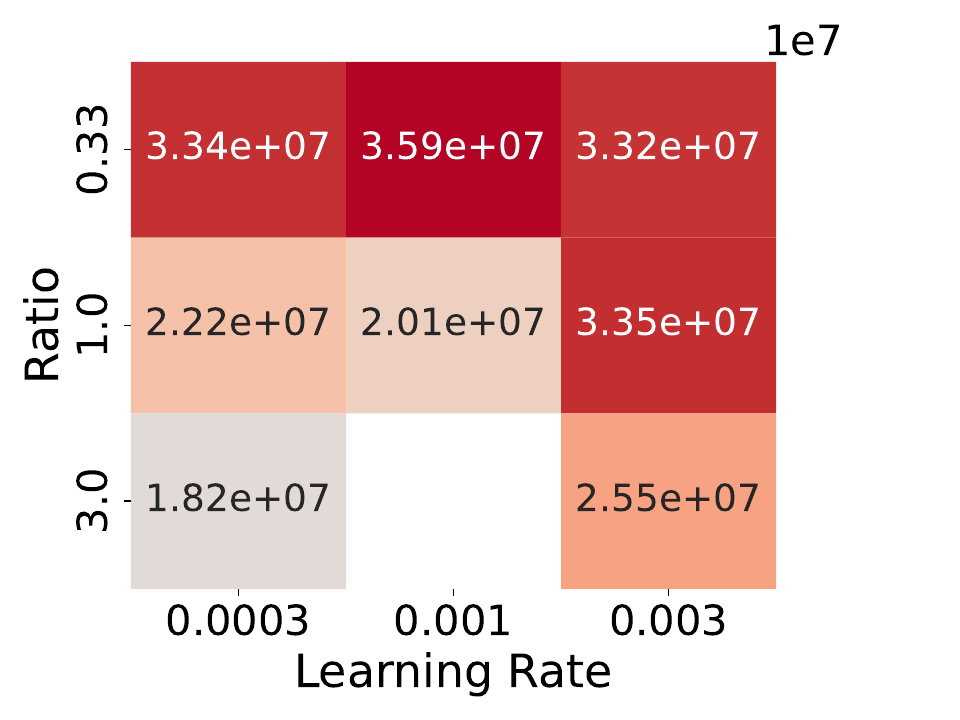 }
  \caption{ PPO, Non-shared Multi-body, Custom entropy }
\end{subfigure}
\hspace{0.1cm}
\begin{subfigure}{0.3\textwidth}
  \centering
  \includegraphics[width=\textwidth]{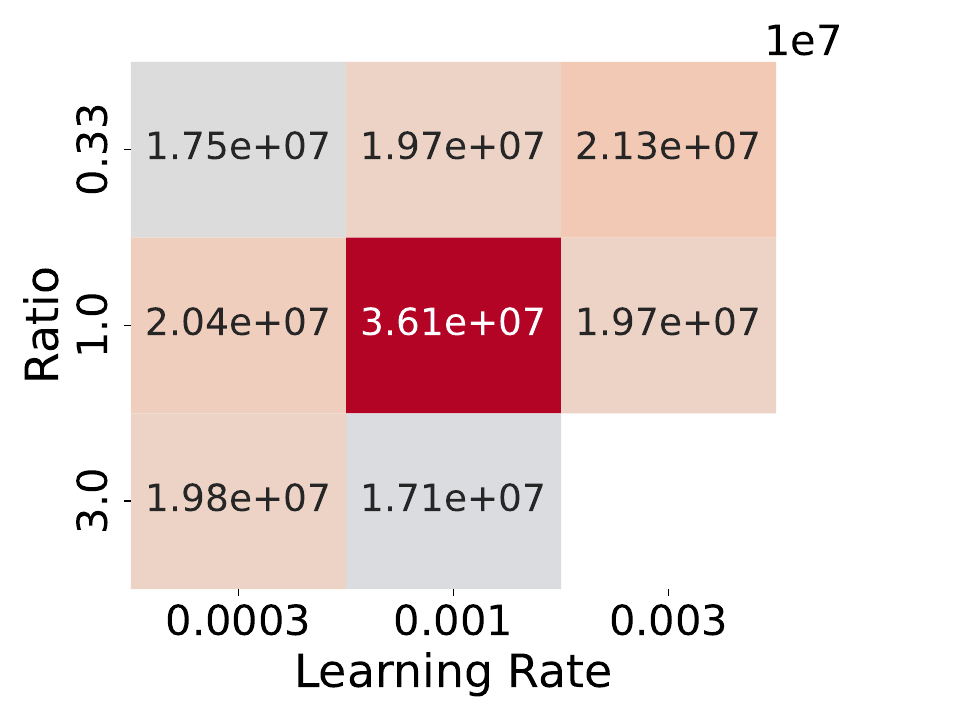 }
  \caption{ PPO, Shared Multi-body, Custom entropy }
\end{subfigure}
\hspace{0.1cm}
\begin{subfigure}{0.3\textwidth}
  \centering
  \includegraphics[width=\textwidth]{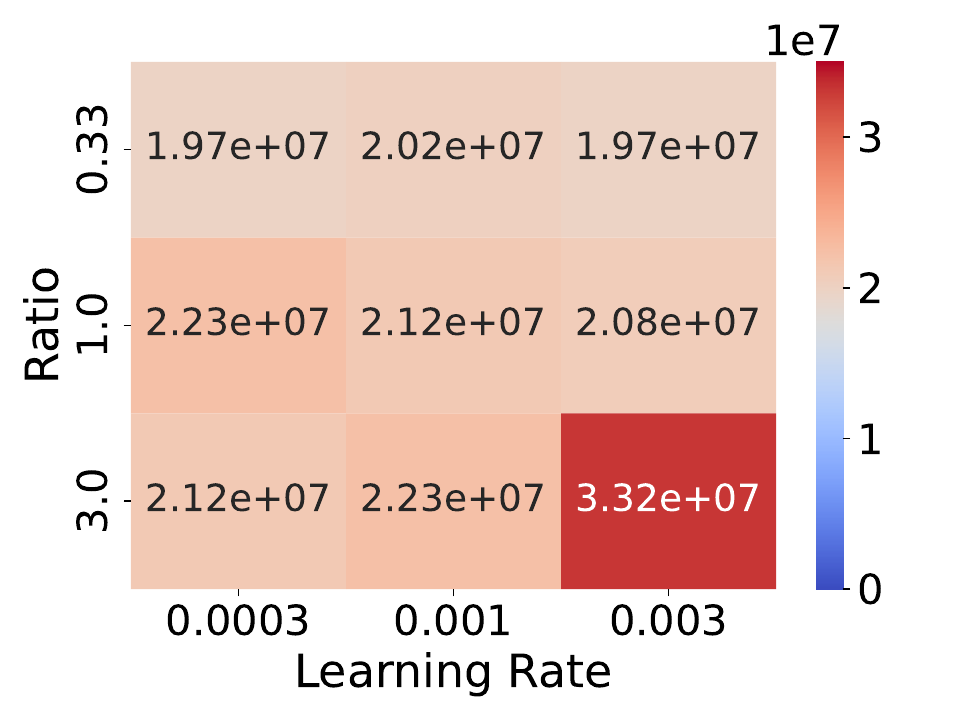 }
  \caption{ PPO, Shared Multi-body, Linear entropy }
\end{subfigure}
\vspace{1em}
\begin{subfigure}{0.3\textwidth}
  \centering
  \includegraphics[width=\textwidth]{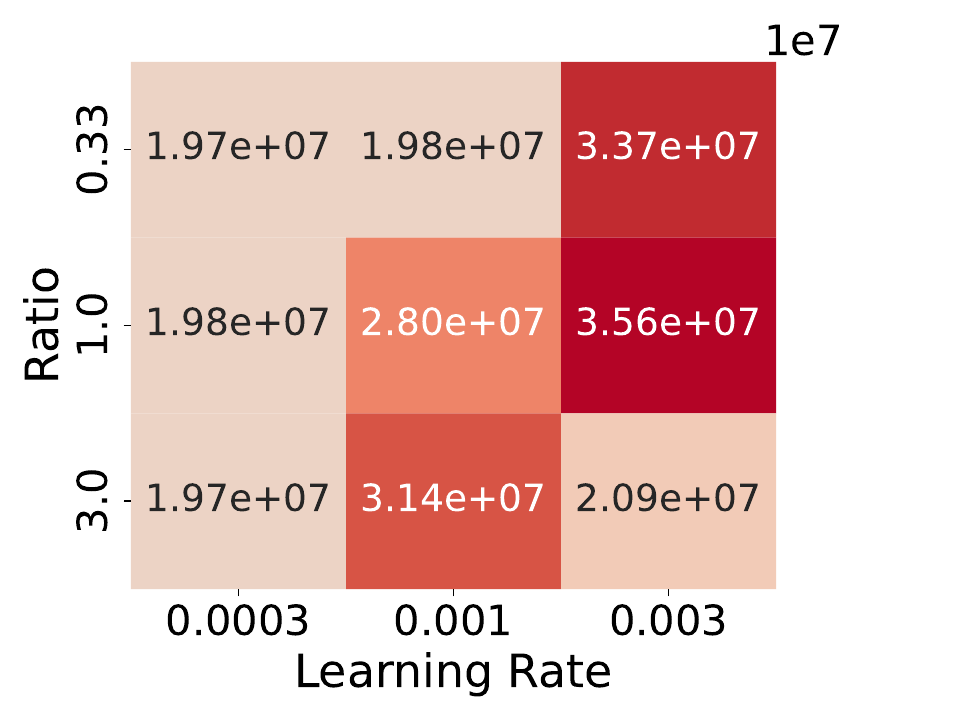 }
  \caption{ PPO, Shared Hypernet, Custom entropy, No PopArt }
\end{subfigure}
\hspace{0.1cm}
\begin{subfigure}{0.3\textwidth}
  \centering
  \includegraphics[width=\textwidth]{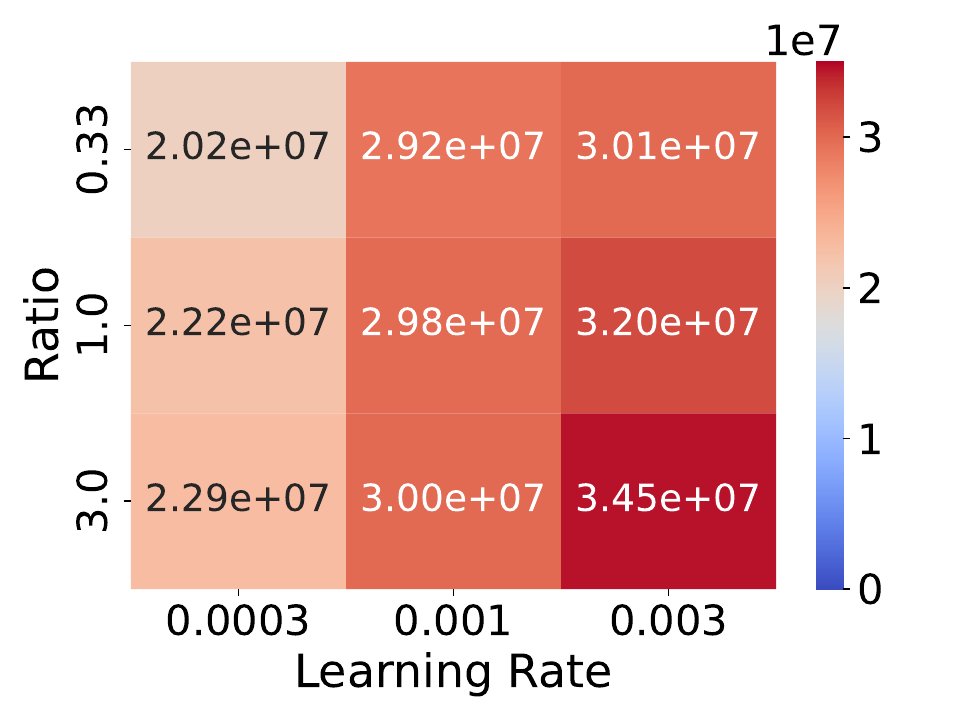 }
  \caption{ PPO, Shared Mult. embed., Custom entropy }
\end{subfigure}
\hspace{0.1cm}
    \caption{Hypervolume of all hyperparameter configurations for our methods on Reacher.}
    \label{fig:hyperparam_our_reacher}
\end{figure}

%%%%%%%%%%%%%%%%%%%%%%%%%%%%%%%%%%%%%%%%%%%%%%%%%%%%%%%%%%%%

% \newpage
\clearpage

\end{document}